\documentclass{article} % For LaTeX2e
\usepackage{iclr2023_conference,times}

\usepackage{amsmath,amsfonts,bm}

\def\eqref#1{equation~\ref{#1}}
\def\1{\bm{1}}

\DeclareMathAlphabet{\mathsfit}{\encodingdefault}{\sfdefault}{m}{sl}
\SetMathAlphabet{\mathsfit}{bold}{\encodingdefault}{\sfdefault}{bx}{n}

\usepackage{hyperref}
\usepackage{url}
\usepackage{multirow}
\usepackage{booktabs}
\usepackage{amssymb}
\usepackage{mathrsfs}
\usepackage{amsmath}
\usepackage{bbding}
\usepackage{graphicx}
\PassOptionsToPackage{table}{xcolor}
\usepackage{colortbl}  %彩色表格需要加载的宏包
\usepackage{wrapfig}
\usepackage{subcaption}
\usepackage{floatrow}
\usepackage{pifont}
\usepackage{color}
\usepackage{anyfontsize}
\usepackage[11pt]{moresize}
\usepackage[export]{adjustbox}
\definecolor{defaultcolor}{gray}{0.9}
\usepackage[linewidth=1pt]{mdframed}

\title{Label Distribution Learning via  Implicit Distribution Representation}

\author{Zhuoran Zheng$^{1}$ and Xiuyi Jia$^{1}$\thanks{Corresponding authors.}\\
	$^{1}$CSE, Nanjing University of Science and Technology
}

\begin{document}

\maketitle

\begin{abstract}
In contrast to multi-label learning, label distribution learning characterizes the polysemy of examples by a label distribution to represent richer semantics.
In the learning process of label distribution, the training data is collected mainly by manual annotation or label enhancement algorithms to generate label distribution.
Unfortunately, the complexity of the manual annotation task or the inaccuracy of the label enhancement algorithm leads to noise and uncertainty in the label distribution training set.
To alleviate this problem, we introduce the implicit distribution in the label distribution learning framework to characterize the uncertainty of each label value.
Specifically, we use deep implicit representation learning to construct a label distribution matrix with Gaussian prior constraints, where each row component corresponds to the distribution estimate of each label value, and this row component is constrained by a prior Gaussian distribution to moderate the noise and uncertainty interference of the label distribution dataset.
Finally, each row component of the label distribution matrix is transformed into a standard label distribution form by using the self-attention algorithm. 
In addition, some approaches with regularization characteristics are conducted in the training phase to improve the performance of the model.

\end{abstract}

\vspace{-5mm}
\section{Introduction}
\vspace{-3mm}
Label distribution learning (LDL) (\cite{geng2016label}) is a novel learning paradigm that characterizes the polysemy of examples. 
In LDL, the relevance of each label to an example is given by an exact numerical value between 0 and 1 (also known as description degree), and the description degree of all labels forms a distribution to fully characterize the polysemy of an example. 
Compared with traditional learning paradigms, LDL is a more generalizable and representational learning paradigm that provides richer semantic information.
\vspace{-1mm}

LDL has been successful in several application domains (\cite{gao2018age,zhao2021robust,chen2021toward,si2022towards}). To obtain the label distribution for learning, there are mainly two ways: one is expert labeling, but labeling is expensive and there is no objective labeling criterion, and the resulting label distribution is highly subjective and ambiguous. 
The other is to convert a multi-label dataset into a label distribution dataset through a label enhancement algorithm (\cite{xu2019label,xu2020variational,zheng2021generalized,zhaofusion}). However, label enhancement lacks a reliable theory to ensure that the label distribution recovered from logical labels converges to the true label distribution, because logical labels provide a very loose solution space for the label distribution, making the solution less stable and less accurate.
\vspace{-1mm}

In summary, the label distribution dataset used for training has a high probability of inaccuracy and uncertainty, which significantly limits the performance of LDL algorithms. To characterize and mitigate the uncertainty of the label distribution, we propose a novel LDL method based on the implicit label distribution representation. 
Our work is inspired by recent work on implicit neural representation in 2D image reconstruction (\cite{sitzmann2020implicit}). 
The key idea of implicit neural representation is to represent an object as a function that maps a sequence of coordinates to the corresponding signal, where the function is de-parameterized by a deep neural network.
%
%In this paper, (从这里开始，后面应该是简要描述下流程，你这里只是说了关键的处理吧，输入是什么？输出的label distribution呢？没头没尾。) the goal of the proposed implicit distribution representation is to generate a label distribution matrix with Gaussian distribution constraints as a customized representation pattern. Specifically, given a coded coordinate, the implicit function takes the coordinate information and queries the local latent codes around the coordinate as inputs, then predicts the label distribution matrix at the given coordinate as an output. （这里最终的输出只写了label distribution matrix，而我们最终的输出是label distribution。要说明白。此外，后面突兀的紧跟着就是latent features around the coordinate，这个latent features是个什么东西，这地方都没有介绍，我为什么要generate latent features?）
%
In this paper, we start with a deep network to extract the latent features of input information.
Then, the latent features are looked up against the encoded coordinate matrix to generate a label distribution matrix (implicit distribution representation).
Finally, the label distribution matrix is computed by a self-attention module to yield a standard label distribution.
Note that the goal of the proposed implicit distribution representation is to generate a label distribution matrix with Gaussian distribution constraints as a customized representation pattern.
%The advantage of implicit neural representation~\cite{sitzmann2020implicit,peng2021neural,shen2022nerp} is the robust modeling  capability for characterizing the continuous representation of an object.
%
\vspace{-1mm}

To efficiently generate latent features around the coordinate,  we design a deep spiking neural network (\cite{yamazaki2022spiking}) with an MLP as an executor to extract latent features in the input information.
The architecture of the whole network consists of multiple layers of linear spiking neurons, and the neurons of different layers conduct a shortcut between them.
Notably, spiking neural networks have two key properties that are different from the representation of artificial neural networks.
First, a standard spiking neural network considers the time characteristics $T$ (taking the image as an example, the input tensor $X \in$ ${\mathbb{R}}^{T \times C \times W \times H}$.) at a multi-step inference mechanism.
Here, we create several pseudo-feature spaces on the native feature space by setting different strategies with data augmentation (\cite{ucar2021subtab}).
These pseudo-features and the native feature are stacked in the time dimension $T$ to achieve the multi-step inference mechanism.
Second, the representation capability of the spiking neural network is underpowered, since the output space is a binarized sequence (\{0 $\cdots$ 1\}).
Therefore, we place a standard MLP in the last layer of the spiking neural network, which projects the features into the real number space.
Our model saves about \textbf{30$\sim$40$\%$} of energy consumption over ANNs with the same network structure on embedded devices such as lynxi HP300, Raspberry Pi, or Apple smartphones (PyTorch 1.2 support M1 Mac).
\vspace{-1mm}

\begin{figure}[t!]
	\centering
	\floatbox[{\capbeside\thisfloatsetup{capbesideposition={right,top},capbesidewidth=5.3cm}}]{figure}[\FBwidth]{
		\caption{\textbf{Our architecture.} This figure shows the architecture of the proposed deep implicit function, which consists of two parts.
	The first part starts with a latent feature prediction stream (SNN with an MLP) that learns the input information to predict the feature maps.
	The second part learns a label distribution matrix to regress a label distribution.
	}  
		\label{fig:task_generalizability}  }{\includegraphics[width=0.6\textwidth]{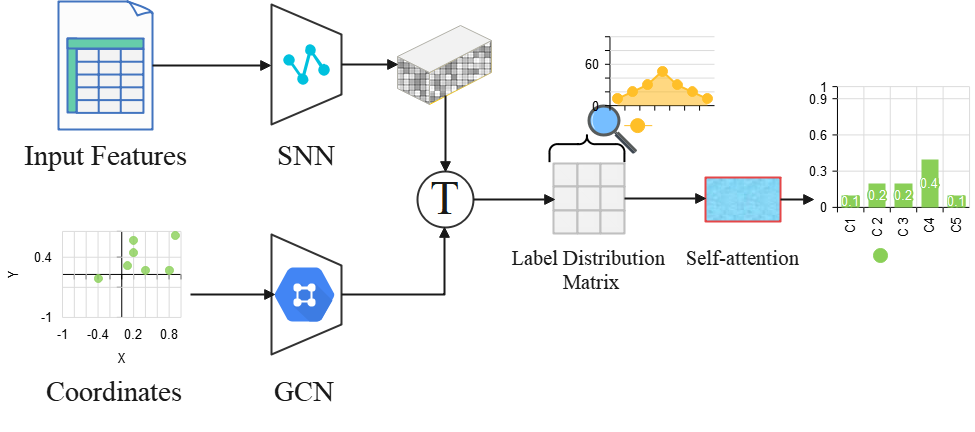}}
		\vspace{-4mm}
\end{figure}

The extraction of latent features provides material for the coordinates to generate the label distribution matrix.
First, the initialized coordinate matrix (the size is $L \times 64$, where $L$ denotes the number of labels, and 64 denotes features of the nodes) is reconstructed by using a GCN.
Note that the features of the nodes meet the Gaussian distribution since the deep network needs to reconstruct the data from a fixed distribution.
Then the coordinate matrix is repeated in $N$ copies, and $N$ denotes the number of samples.
%The edges between nodes are provided by the Gram matrix computed in the label space.
%
Next, the coordinate matrix computes a label distribution matrix (the size is $N \times L \times 2L$) in the latent feature space by looking up the table\footnote[1]{\textcolor{blue}{\tiny \url{https://pytorch.org/tutorials/intermediate/spatial_transformer_tutorial.html}}}.
Each component of the label distribution matrix represents the distribution of each label value, and the components are constrained by a priori Gaussian distributions.
Finally, the label distribution matrix leverages a self-attention mechanism to obtain the corresponding label distribution of the samples.
\vspace{-1mm}

%So far, a tricky problem to face is that the dataset for label distribution learning is a table-based format.
%
Yet, deep learning-based approaches are prone to overfitting manually extracted features.
To alleviate the problem, we propose some regularization approaches to boost the performance of the model, and a new dataset based on the image comprehension task is released.
\textbf{Our contribution includes:} \textbf{(i)} For LDL, this is a novel method to obtain the label distribution of a sample through the implicit distribution representation.
\textbf{(ii)} Spiking neural network with an MLP is developed to save energy consumption of mobile devices, and correlations between labels are deeply mined by a graph convolutional network.
\textbf{(iii)} Facing the LDL task, some regularization techniques are designed to boost the performance of the model and a new LDL dataset is released.
\vspace{-5mm}
\section{Proposed Method}
\vspace{-4mm}
As shown in Figure~\ref{fig:task_generalizability}, our architecture is a two-stage scheme, which is based on an implicit representation approach.
In the first stage (\textbf{latent feature extraction}), the raw features $\mathcal{X}$ are fed into the encoder for encoding to construct a latent feature space $F_{f} \in$ ${\mathbb{R}}^{L \times H \times W}$.
%
%Notably, in the datasets released by the label distribution learning, the raw feature space size is inconsistent, so $H$ and $W$ are not fixed (the variables ensure that they are a power of 2).
%
In the second stage (\textbf{label distribution matrix learning}), the coordinate tensor (coordinate matrix is reshaped after GCN) $C_{r} \in$ ${\mathbb{R}}^{L \times L \times 2)}$ is looked up in the latent feature space $F_{f}$ to generate a label distribution matrix $\mathcal{M} \in$ ${\mathbb{R}}^{L \times 2L}$, and finally a label distribution $\mathcal{D}$ is obtained by using self-attention algorithm.
Furthermore, some regularization terms are introduced to our model to boost its performance.
%

%\textbf{Preliminaries.}
%
%Let $\mathcal{X}=\mathbb{R}^{m}$ denote the $m$-dimensional feature space, and $\mathcal{D}=\left\{d^{y_{1}}, d^{y_{2}}, \ldots, d^{y_{c}}\right\}$ denote the label space, where $c$ is the number of labels and $\mathcal{Y}=$ $\left\{y_{1}, y_{2}, \ldots, y_{c}\right\}$ is the finite label set.
%
%Given a training set $T=\left\{\left(\mathbf{x}_{1}, \mathbf{d}_{1}\right),\left(\mathbf{x}_{2}, \mathbf{d}_{2}\right), \ldots,\left(\mathbf{x}_{n}, \mathbf{d}_{n}\right)\right\}$, where $\mathbf{x}_{i}=$ $\left\{x_{i}^{(1)}, x_{i}^{(2)}, \ldots, x_{i}^{(m)}\right\} \in \mathcal{X}$ is the feature vector of the $i$-th instance and $\mathbf{d}_{i}=\left\{d_{i}^{y_{1}}, d_{i}^{y_{2}}, \ldots, d_{i}^{y_{c}}\right\}$ is the label distribution associated with $\mathbf{x}_{i} . d_{i}^{y_{j}}$ is called the description degree of label $y_{j}$ with respect to $\mathbf{x}_{i}$, and satisfies $\sum_{j} d_{x_{i}}^{y_{j}}=1, d_{x_{i}}^{y_{j}} \geq 0$. 
%
%LDL aims to learn a mapping $F: \mathcal{X} \rightarrow \mathcal{D}$ from the training set $T$ for predicting the corresponding label distribution of input instances.

\textbf{Latent Feature Extraction.}
Our model starts with a latent feature extraction task, which is based on a spiking neural network.
Spiking neural networks have a high potential value as a 3rd deep model and are efficient and interpretable.
In the LDL task, we develop a non-fixed \textcolor{blue}{Network Architecture} and a \textcolor{blue}{Network Implementation} due to the diversity of LDL dataset formats.

\textbf{\textcolor{blue}{Network Architecture:}}
Our network consists of 17 layers of function units, which include 8 linear units, 8 nonlinear units, and a transformation layer (including an MLP, a mean operator, and a reshape operator).
Essentially, this is a residual network and the transformation layer at the tail of the network.
Except for the first layer (the number of neurons is the dimensionality of the input features), each linear unit contains 1024 neurons, followed closely by a ReLU non-linearity, and the last layer is a reshape function to generate the feature space $F_{f}$.
On datasets with a smaller feature space $\mathcal{X}$ (e.g., Gene dataset has only 36 features), each linear layer includes 64 neurons.
We also attempted to use other activation functions than ReLU, such as PReLU, Swish, and Sigmoid, but without any advantage.
%
%Note that since the representational ability of the spiking neural network is limited, we employ a standard MLP in the transformation layer to boost the performance of the model.
%
Besides, since the spiking network involves a time dimension, the time dimension $T$ is squeezed in the output space of the spiking network by using a mean operator.
The network's tailgate should notice that the feature map is reshaped as $F_{f}$ using a reshape operator and the MLP (the number of neurons is $L \times W \times H$.), where $H$ and $W$ are 32.
%
%In the Gene dataset, $H$ and $W$ are 8.
%
\vspace{-2mm}

\textbf{\textcolor{blue}{Network Implementations:}}
The simplified residual network is implemented as an ANN in two frameworks: PyTorch 1.12 (a standard deep network training library) and SpikingJelly 0.0.13 (a library for deep learning applications that is part of the PyTorch ecosystem).
The SpikingJelly implementation can be run in spiking or non-spiking modes.
To train the spiking network, we use an ANN-SNN conversion training approach.
The conversion is simple, we convert the trained model on PyTorch with the help of an ann2snn.Converter (SpikingJelly) to obtain an SNN model.
Note that the conversion method defines six modes (max, $99.9\%$, $1.0 \backslash 2$, $1.0 \backslash 3$, $1.0 \backslash 4$, $1.0 \backslash 5$) to obtain SNN with different accuracy, and in this paper, we chose $99.9\%$.
The literature (\cite{patel2021spiking}) provides a theoretical basis for analyzing the conversion of ANN to SNN.
%

%zhao2022continuous
\textbf{Label Distribution Matrix Learning.}
Inspired by confident learning (\cite{northcutt2021confident}), we develop a matrix of label distribution with a Gaussian before estimating the uncertainty of the labels.
For the data distribution of each label value, we use a Gaussian distribution to delimit the distribution rather than other multi-peaked distribution priors, and numerous kinds of literature have verified that this approach can eliminate the uncertainty (\cite{liu2021ngdnet,zheng2021uncertainty,ghosh2021uncertainty,li2022unimodal}).
Furthermore, to capture the global correlation between labels to generate a standard label distribution, we employ a self-attention mechanism to model the label distribution matrix.
To obtain a label distribution matrix $\mathcal{M}$, we need a latent feature $F_{f}$ from the SNN and a coordinate matrix $C_{b}:\text{GCN} (C_{r})$ (GCN denotes the graph convolution network) that passes through the graph convolution network.
Specifically, first, we initialize a matrix of coordinates $C_{r} \in$ ${\mathbb{R}}^{L \times 64}$ based on the functions (torch.randn) provided by PyTorch.
$L$ denotes the number of nodes and $64$ denotes that we assign one feature vector to each node, with each feature vector sampled in a Gaussian distribution.
To build a graph structure the data needs information about the edges, where there is an edge with no direction between the nodes.
So far, we built the node with the edge information and have the ability to integrate it into a graph to input into the GCN.
Our GCN includes four graph convolution layers and four activation layers, where the activation function uses ReLU.
These four graph convolutions include 64, 128, 256, and $L \times L \times 2$ neurons respectively.
There is one key message to note, the output matrix $C_{b}$ of the GCN is filled (torch.repeat) with the same number of samples as the latent feature space $F_{f}$.
Then, we use $C_{b}$ to look up the table in $F_{f}$ to obtain a matrix by using flatten operations.
The $L$ label distribution values include a $1 \times 2L$ vector to form a label distribution matrix $\mathcal{M}$.
Each vector is constrained by a designated Gaussian distribution $\bar{\mathcal{M}}$ with parameters whose mean is the value of the label distribution and variance of 0.5.
Finally, this matrix is squeezed by using a self-attention algorithm (\cite{vaswani2017attention}) to obtain the corresponding label distribution for the samples.

\textbf{Regularization Techniques.}
We propose two techniques (\textcolor{blue}{Linear Normalization Function}, \textcolor{blue}{Data Augmentation}) to boost the performance of our model.

\textbf{\textcolor{blue}{Linear Normalization Function:}} 
Currently, most existing LDL algorithms use Softmax to output a vector in the tail of the model.
\begin{equation}
\text{Softmax}(d_{i}^{y_{i}}) =\frac{\exp^{d_{i}^{y_{i}}}}{\sum_{c=1}^{C} \exp^{d_{c}^{y_{c}}}},
\end{equation}
where $d_{i}^{y_{i}}$ denotes a label value in the vector and $C$ denotes the number of elements.
%
%Linear normalization methods are widely considered, such as cosFormer~\cite{?}, LineFormer~\cite{?}, and these methods improve the efficiency of the algorithm.
%
Since the complexity of exponential operations is much higher than linear operations, especially in the training phase of the model (e.g. Gamma correction (\cite{ju2019idgcp})).
Several algorithms (such as cosFormer (\cite{qin2022cosformer})) seek to replace exponential operations.
Here, we develop a linear function Lnf to replace Softmax to output a label distribution vector. 
%Based on the outstanding performance of the linear normalization method (such as cosFormer~\cite{?}, LineFormer~\cite{?}), we design a linear operator to replace it to boost the performance of the model.
%
Then predicted label distribution $\widehat{\mathbf{d}_{i}}=$ $\left\{\widehat{d_{i}^{y_{1}}}, \widehat{d_{i}^{y_{2}}}, \ldots, \widehat{d_{i}^{y_{c}}}\right\}$ can be obtained by:
\begin{equation}
\text{Lnf}(\widehat{d_{i}^{y_{i}}}) =\frac{d_{i}^{y_{i}} + |\mathcal{D}_{\text{min}}|}{\sum_{c=1}^{C} (d_{c}^{y_{c}} + |\mathcal{D}_{\text{min}}|)},
\end{equation}
where $|\mathcal{D}_{\text{min}}|$ denotes the absolute value of the minimum of the predicted label distribution values.
Essentially, our method offsets the values of the label distribution to the positive domain of the x-axis.
We evaluate both Softmax and linear normalization function on 12 datasets, and our network with the linear normalization function converges \colorbox{defaultcolor}{2.1 $\times$} faster than the model with Softmax.
%
%By a power base modeling approach is easily disturbed on a noisy dataset.
%
%However, the linear computational error is also linear in the gap.
%

\textbf{\textcolor{blue}{Data Augmentation}}:
Most LDL datasets are difficult to augment the samples with expert knowledge since the input features are extracted manually.
To overcome this challenge, we introduce a simple and data-agnostic data augmentation routine, termed mixup (\cite{zhang2017mixup}). In a nutshell, mixup constructs virtual training examples:
\begin{equation}
\begin{aligned}
&\tilde{x}=\lambda x_{i}+(1-\lambda) x_{j}, \quad \text { where } x_{i}, x_{j} \text { are raw input vectors } \\
&\tilde{y}=\lambda y_{i}+(1-\lambda) y_{j}, \quad \text { where } y_{i}, y_{j} \text { are label distribution vectors }
\end{aligned}
\end{equation}
$\left(x_{i}, y_{i}\right)$ and $\left(x_{j}, y_{j}\right)$ are two examples drawn at random from our training data, and $\lambda \in[0,1]$.
However, a deep network may still be over-fitted during the training phase, which leads to poor generalization.
We use the random masking scheme endowed on mixup, formally expressed as:
\begin{equation}
\begin{aligned}
&\tilde{x}=\lambda x_{i} \times \text{mask} +(1-\lambda) x_{j} \times \text{mask}, \quad \text { where } x_{i}, x_{j} \text { are raw input vectors } \\
&\tilde{y}=\text{Lnf}(\lambda y_{i}+(1-\lambda) y_{j}), \quad \text { where } y_{i}, y_{j} \text { are label distribution vectors }
\end{aligned}
\end{equation}
the mask is a vector represented by 1 or 0. In this paper, a mask contains 80\% of the scalar 1 and the rest is 0.
Moreover, to solidify the definition of the label distribution, we use Lnf to normalize the synthesized label vectors.
We also tried using other regularization schemes, such as random cropping, and the results do not improve significantly.

{\flushleft \textbf{Loss Function.}}
We optimize the weights and biases of the proposed network by minimizing the $L_{1}$, KullbackLeibler (K-L) divergence, and regularization of the label distribution matrix on the training set,
\begin{equation}
\mathcal{L}=\frac{1}{N}\sum_{i=1}^{N} \left \| \widehat{d_{i}}-d_{i} \right \| + \lambda\mathcal{L}_{kl} + \beta\sum_{i=1}^{N}\sum_{i=1}^{L}\mathcal{L}_{2}(\mathcal{M}_{ii}, \bar{\mathcal{M}_{ii}}),
\end{equation}
where $N$ is the number of training samples, $d_{i}$ is the label distribution vector by our model, and $\widehat{d}$ is the corresponding ground truth. The weight $\lambda$ of the loss term $\mathcal{L}_{kl}$ (K-L) is set to 0.01 in our experiments.
In addition, the weight $\beta$ of the loss term (label distribution matrix) is set to 0.1 in our experiments. 
Since it is difficult to capture useful information by encoding a small number of tokens, encoding for sequences of long tokens can extract higher-order semantics, such as perceptual loss (\cite{johnson2016perceptual}).
Therefor, we used the above-mentioned learning strategy on the datasets with several labels less than 20. 
On the datasets with the number of labels greater than 20, we used $L_{1}$, K-L divergence, perceptual loss, and regularization of the label distribution matrix on the training set,
\begin{equation}
\mathcal{L}=\frac{1}{N}\sum_{i=1}^{N} \left \| \widehat{d_{i}}-d_{i} \right \| + \lambda_{1}\mathcal{L}_{kl} + \lambda_{2}\mathcal{L}_{p} + \beta\sum_{i=1}^{N}\sum_{i=1}^{L}\mathcal{L}_{2}(\mathcal{M}_{ii}, \bar{\mathcal{M}_{ii}}),
\end{equation}
we use MLPs as the pre-trained model for the perceptual loss function ($\mathcal{L}_{p}$).
The weight $\lambda_{1}$, $\lambda_{2}$, and $\beta$ are set to 0.01, 0.01, 0.08 in our experiments, respectively. 
%$\mathcal{L}_{p}$ is used to improve visual quality.
%We use VGG19~\cite{SimonyanZ14a} as the pre-trained model for the perceptual loss function.
%We also try to use the total variation and adversarial losses, but we notice that
%the $L_{1}$ and $\mathcal{L}_{p}$ can generate vivid colors and clear texture in the deblurred results.
%
These MLPs consist of three linear layers with fixed neurons and three activation layers (ReLU), where the number of neurons is the same as the number of labels, using Kaiming initialization (\cite{he2015delving}).
\vspace{-4mm}
\section{Experiments}
\vspace{-3mm}
{\flushleft \textbf{Algorithm Configurations.}}
We conduct experiments on 12 datasets, the characteristics of the datasets are reported in Table~\ref{T1}.
%
%All datasets are configured by reference to ~\cite{?}. 
Except for dataset wc-LDL, the configuration of all other datasets is referenced to (\cite{wang2021label}).
%We also add the Yeast dataset and a newly collected dataset.
%
This new release dataset (wc-LDL) has 500 watercolor images and corresponding label distribution (12 emotions).
wc-LDL is constructed with a thorough discussion in the supplementary material.
%The label has 12 emotions in total, and each pair of images is evaluated by 10 professionals.
%
%As a result of the final evaluation, a mean value is taken as the distribution value of the label space.
We develop the SNN with configurations on different datasets also summarized in Table~\ref{T1}.
To evaluate the performance of LDL models, we use the six metrics proposed by (\cite{geng2016label}), including Chebyshev distance $\downarrow$, Clark distance $\downarrow$, Canberra distance $\downarrow$, KL divergence $\downarrow$, Cosine similarity $\uparrow$, and Intersection similarity $\uparrow$.
LF and DA denote the loss function and data augmentation method, respectively.
$\downarrow$ represents the indicator's performance favoring low values and $\uparrow$ represents the indicator's performance favoring high values.
$\mathscr{L}_{1,2:end}$ denotes the number of neurons in the head layer and the rest of the layers of the SNN. 
% $\XSolidBrush$
\begin{table*}[!ht] \tiny
	\begin{center}
		\vspace{-1mm}
		\caption{Statistics of the experimental datasets with models.}
		\vspace{-0mm}
		\label{T1}
		\begin{tabular}{llccclcc}
			\toprule
			ID & Dataset 	          & Examples	& Features  &Labels   &Number of neurons of the SNN    &LF  & DA          \\ 
			\midrule
			1  &  wc-LDL   &  500        & 243       & 12       &$\mathscr{L}_{1,2:end}$ $\rightarrow$ [243, 1024]  & Eq.5 & \Checkmark \\
			2  &  SJAFFE          &  213        & 243       & 6        &$\mathscr{L}_{1,2:end}$ $\rightarrow$ [243, 1024]  & Eq.5 &  \Checkmark \\
			3  &  SBU-3DFE        &  2500       & 243       & 6       &$\mathscr{L}_{1,2:end}$ $\rightarrow$ [243, 1024]  & Eq.5 & \XSolidBrush \\
			4  &  Scene           &  2000       & 294       & 9       &$\mathscr{L}_{1,2:end}$ $\rightarrow$ [294, 1024]  & Eq.5 & \XSolidBrush\\
			5  &  Gene            &  17892      & 36        & 68      &$\mathscr{L}_{1,2:end}$ $\rightarrow$ [36, 64]  & Eq.6 & \XSolidBrush\\
			6  &  Movie           &  7755       & 1869      & 5       &$\mathscr{L}_{1,2:end}$ $\rightarrow$ [1869, 1024]  & Eq.5 & \XSolidBrush\\
			7  &  M2B             &  1240       & 250       & 5       &$\mathscr{L}_{1,2:end}$ $\rightarrow$ [256, 1024]  & Eq.5 &  \Checkmark \\
			8  &  SCUT            &  1500       & 300       & 5      &$\mathscr{L}_{1,2:end}$ $\rightarrow$ [300, 1024]  & Eq.5 &  \Checkmark \\
			9  &  fbp5500         &  5500       & 512       & 5       &$\mathscr{L}_{1,2:end}$ $\rightarrow$ [512, 1024]  & Eq.5 & \XSolidBrush\\
			10  &  RAF-ML         &  4908       & 200       & 6       &$\mathscr{L}_{1,2:end}$ $\rightarrow$ [200, 1024]  & Eq.5 & \XSolidBrush\\
			11  &  Twitter        &  10040      & 200       & 8      &$\mathscr{L}_{1,2:end}$ $\rightarrow$ [200, 1024]  & Eq.5 & \XSolidBrush\\
			12  &  Flickr         &  11150      & 200       & 8       &$\mathscr{L}_{1,2:end}$ $\rightarrow$ [200, 1024]  & Eq.5 & \XSolidBrush\\
		%	13  &  Alpha         &  2465       & 24       & 18       &$\mathscr{L}_{1,2:end}$ $\rightarrow$ [24, 1024]  & \textbf{\textcolor{blue}{Eq.6}} &  \Checkmark \\
		%	14  &  Cdc           &  2465      & 24       & 15      &$\mathscr{L}_{1,2:end}$ $\rightarrow$ [24, 64]  & \textbf{\textcolor{blue}{Eq.6}} &  \Checkmark \\
		%	15  &  Cold          &  2465      & 24       & 4       &$\mathscr{L}_{1,2:end}$ $\rightarrow$ [24, 64]  & Eq.5 &  \Checkmark \\
		%	16  &  Diau          &  2465       & 24       & 7       &$\mathscr{L}_{1,2:end}$ $\rightarrow$ [24, 64]  & Eq.5 &  \Checkmark \\
		%	17  &  Dtt           &  2465      & 24       & 4      &$\mathscr{L}_{1,2:end}$ $\rightarrow$ [24, 64]  & Eq.5 &  \Checkmark \\
		%	18  &  Elu           &  2465      & 24       & 14       &$\mathscr{L}_{1,2:end}$ $\rightarrow$ [24, 64]  & Eq.5 &  \Checkmark \\
		%	19  &  Heat          &  2465       & 24       & 6       &$\mathscr{L}_{1,2:end}$ $\rightarrow$ [24, 64]  & Eq.5 &  \Checkmark \\
		%	20  &  Spo           &  2465      & 24        & 6      &$\mathscr{L}_{1,2:end}$ $\rightarrow$ [24, 64]  & Eq.5 &  \Checkmark \\
		%	21  &  Spo5          &  2465      & 24       & 3       &$\mathscr{L}_{1,2:end}$ $\rightarrow$ [24, 64]  & Eq.5 &  \Checkmark \\
		%	22  &  Spoem         &  2465      & 24       & 2       &$\mathscr{L}_{1,2:end}$ $\rightarrow$ [24, 64]  & Eq.5 &  \Checkmark \\
			\bottomrule
		\end{tabular}%
		\vspace{-4mm}
	\end{center}
\end{table*}

\vspace{-5mm}
{\flushleft \textbf{Experimental Setting.}}
We conduct comparative experiments with five LDL algorithms (BFGS-LLD (\cite{geng2016label}), LDL-LRR (\cite{jia2021label}), LDL-LCLR (\cite{ren2019label}), LDLSF (\cite{ren2019labelf}) and LALOT (\cite{zhao2018label})) which also used data augmentation schemes on 12 datasets.
BFGS-LLD is based on a linear model, the loss function is K-L divergence, and the optimization method is the quasi-Newton approach.
LDL-LRR and LDL-LCLR both consider label correlations in the learning process, with the former considering the order relationship of the labels and the latter capturing global relationships between labels.
For LDL-LRR, the parameters $\lambda$ and $\beta$ are selected from $10^{\{-6,-5, \ldots,-2,-1\}}$ and $10^{\{-3,-2, \ldots, 1,2\}}$, respectively. For LDL-LCLR, the parameters $\lambda_{1}, \lambda_{2}, \lambda_{3}, \lambda_{4}$ and $k$ are set to $0.0001,0.001,0.001,0.001$ and 4, respectively. 
LDLSF leverages label-specific features and common features simultaneously, whose parameters $\lambda_{1}, \lambda_{2}$ and $\lambda_{3}$ are selected from $10^{\{-6,-5, \ldots,-2,-1\}}$, respectively, and $\rho$ is set to $10^{-3}$. 
LALOT adopts optimal transport distance as the loss function, the trade-off parameter $C$ and the regularization coefficient $\lambda$ is set to $200$ and $0.2$, respectively.
Our approach for experimental settings is reported in Table~\ref{T2}.
It is worth noting that Early stopping and Greed soup (\cite{wortsman2022model}) are also used on all the datasets where the comparison algorithm is executed.

\begin{table*}[!ht] \tiny
	\begin{center}
		\vspace{-0mm}
		\caption{Training configuration of our model.}
		\vspace{-0mm}
		\label{T2}
		\begin{tabular}{llcccccc}
			\toprule
			ID & Dataset 	        & Batch size  & Epoch  &Learning rate   &Weight decay    &Early stopping  &Greed soup          \\ 
			\midrule
			1  &  wc-LDL   &  500        & 200       & $2 \times 10^{-3}$       &$1 \times 10^{-4}$  & \Checkmark & \Checkmark \\
			2  &  SJAFFE            &  213        & 200       & $2 \times 10^{-2}$        &$1 \times 10^{-4}$   & \Checkmark &  \Checkmark \\
			3  &  SBU-3DFE          &  1000       & 200       & $1 \times 10^{-3}$       &$1 \times 10^{-4}$    & \Checkmark & \Checkmark \\
			4  &  Scene             &  1000       & 120       & $1 \times 10^{-3}$       &$1 \times 10^{-4}$    & \Checkmark & \Checkmark\\
			5  &  Gene              &  5000       & 150       & $2 \times 10^{-3}$      &$1 \times 10^{-4}$   & \Checkmark & \Checkmark\\
			6  &  Movie             &  2000       & 100      & $2 \times 10^{-3}$       &$1 \times 10^{-4}$    & \Checkmark & \Checkmark\\
			7  &  M2B               &  500        & 150       & $2 \times 10^{-3}$       &$1 \times 10^{-4}$    & \Checkmark &  \Checkmark \\
			8  &  SCUT              &  500        & 150       & $1 \times 10^{-2}$      &$1 \times 10^{-4}$   & \Checkmark &  \Checkmark \\
			9  &  fbp5500           &  1500       & 300       & $2 \times 10^{-2}$       &$1 \times 10^{-4}$    & \Checkmark & \Checkmark\\
			10  &  RAF-ML           &  2000       & 100       & $1 \times 10^{-3}$       &$1 \times 10^{-4}$    & \Checkmark & \Checkmark\\
			11  &  Twitter          &  5000      & 200       & $1 \times 10^{-3}$      &$1 \times 10^{-4}$    & \XSolidBrush & \XSolidBrush\\
			12  &  Flickr           &  5000      & 200       & $1 \times 10^{-2}$       &$1 \times 10^{-4}$    & \Checkmark & \Checkmark\\
		%	13  &  Alpha            &  1200       & 100       & $1 \times 10^{-2}$       &$2 \times 10^{-5}$    & \XSolidBrush &  \XSolidBrush \\
		%	14  &  Cdc              &  1200      & 100       & $1 \times 10^{-2}$      &$2 \times 10^{-5}$    & \XSolidBrush &  \XSolidBrush \\
		%	15  &  Cold             &  1200      & 100       & $1 \times 10^{-2}$       &$2 \times 10^{-5}$    & \XSolidBrush &  \XSolidBrush \\
		%	16  &  Diau             &  1200      & 100       & $1 \times 10^{-2}$       &$2 \times 10^{-5}$   & \XSolidBrush &  \XSolidBrush \\
		%	17  &  Dtt              &  1200      & 100       & $1 \times 10^{-2}$      &$2 \times 10^{-5}$  & \XSolidBrush &  \XSolidBrush \\
		%	18  &  Elu              &  1200      & 100       & $1 \times 10^{-2}$       &$2 \times 10^{-5}$  & \XSolidBrush &  \XSolidBrush \\
		%	19  &  Heat             &  1200      & 100       & $1 \times 10^{-2}$       &$2 \times 10^{-5}$    & \XSolidBrush &  \XSolidBrush \\
		%	20  &  Spo              &  1200      & 100        & $1 \times 10^{-2}$      &$2 \times 10^{-5}$   & \XSolidBrush &  \XSolidBrush \\
		%	21  &  Spo5             &  1200      & 100       & $1 \times 10^{-2}$       &$2 \times 10^{-5}$    & \XSolidBrush &  \XSolidBrush \\
		%	22  &  Spoem            &  1200      & 100       & $1 \times 10^{-2}$       &$2 \times 10^{-5}$   & \XSolidBrush &  \XSolidBrush \\
			\bottomrule
		\end{tabular}%
		\vspace{-2mm}
	\end{center}
\end{table*}

\begin{table*}[!htb] \tiny
	\begin{center}
		\vspace{-0mm}
		\caption{The performance of our proposed method with the comparison algorithms on 12 datasets. All algorithms enforce techniques such as early stopping, data augmentation, and greedy-soup to evaluate 12 datasets on the GPU. Our method is marked as \colorbox{defaultcolor}{gray}.}
			%, with the last column (BLB) showing the degree of performance improvement for each algorithm compared to those without regularization techniques.}
		\vspace{-0mm}
		\label{T3}
		%\resizebox{\linewidth}{!}{
			\begin{tabular}{c|c|cccccc}
				\toprule
				 Dataset                             & Algorithm	   & Chebyshev $\downarrow$  & Clark $\downarrow$   &Canberra $\downarrow$   &K-L $\downarrow$    &Cosine $\uparrow$   &Intersection $\uparrow$           \\ 	
				\midrule 
				\rowcolor{defaultcolor}                         
				&     Ours     & 0.0779 $\pm$ \textbf{\fontsize{5}{6} \selectfont 0.0021}           & 0.3980 $\pm$ \textbf{\fontsize{5}{6} \selectfont0.0051}       &0.7779 $\pm$ \textbf{\fontsize{5}{6} \selectfont0.0030}          & 0.04040 $\pm$ \textbf{\fontsize{5}{6} \selectfont0.0020}     & 0.9883 $\pm$ \textbf{\fontsize{5}{6} \selectfont0.0009}      & 0.8778 $\pm$ \textbf{\fontsize{5}{6} \selectfont0.0014}                \\
				
				&  LDL-LRR     & 0.1122 $\pm$ \textbf{\fontsize{5}{6} \selectfont0.0030}          & 0.4772 $\pm$ \textbf{\fontsize{5}{6} \selectfont0.0036}      &0.8802 $\pm$ \textbf{\fontsize{5}{6} \selectfont0.0024}          & 0.05533 $\pm$ \textbf{\fontsize{5}{6} \selectfont0.0049}     & 0.9510 $\pm$ \textbf{\fontsize{5}{6} \selectfont0.0022}      & 0.8555 $\pm$ \textbf{\fontsize{5}{6} \selectfont0.0047}                   \\
				
				&  LDL-LCLR    & 0.1057 $\pm$ \textbf{\fontsize{5}{6} \selectfont 0.0019}           & 1.0569 $\pm$ \textbf{\fontsize{5}{6} \selectfont 0.0039}       &0.7890 $\pm$ \textbf{\fontsize{5}{6} \selectfont 0.0039}          & 0.05045 $\pm$ \textbf{\fontsize{5}{6} \selectfont 0.0037}     & 0.9668 $\pm$ \textbf{\fontsize{5}{6} \selectfont 0.0049}      & 0.8383 $\pm$ \textbf{\fontsize{5}{6} \selectfont 0.0018}                     \\
				
				&  LDLSF       & 0.1009 $\pm$ \textbf{\fontsize{5}{6} \selectfont0.0038}           & 0.4199 $\pm$ \textbf{\fontsize{5}{6} \selectfont0.0044}       &0.9008 $\pm$ \textbf{\fontsize{5}{6} \selectfont0.0015}          & 0.05199 $\pm$ \textbf{\fontsize{5}{6} \selectfont0.0040}     & 0.9779 $\pm$ \textbf{\fontsize{5}{6} \selectfont0.0018}      & 0.8660 $\pm$ \textbf{\fontsize{5}{6} \selectfont0.0022}                     \\
				
				&  LALOT       & 0.0989 $\pm$ \textbf{\fontsize{5}{6} \selectfont0.0019}          & 0.6689 $\pm$ \textbf{\fontsize{5}{6} \selectfont0.0019}       &0.8089 $\pm$ \textbf{\fontsize{5}{6} \selectfont0.0049}          & 0.04778 $\pm$ \textbf{\fontsize{5}{6} \selectfont0.0018}     & 0.9476 $\pm$ \textbf{\fontsize{5}{6} \selectfont0.0020}      & 0.8700 $\pm$ \textbf{\fontsize{5}{6} \selectfont0.0033}                    \\
				
				\multirow{-6}{*}{wc-LDL} 		            &  BFGS-LLD    & 0.1229 $\pm$ \textbf{\fontsize{5}{6} \selectfont0.0039}           & 1.5657 $\pm$ \textbf{\fontsize{5}{6} \selectfont0.0021}       &0.7998 $\pm$ \textbf{\fontsize{5}{6} \selectfont0.0020}          & 0.04998 $\pm$ \textbf{\fontsize{5}{6} \selectfont0.0051}     & 0.9704 $\pm$ \textbf{\fontsize{5}{6} \selectfont0.0036}      & 0.8611 $\pm$ \textbf{\fontsize{5}{6} \selectfont0.0016}                 \\
				
				\midrule
				\rowcolor{defaultcolor} 
				&  Ours        & 0.0854 $\pm$ \textbf{\fontsize{5}{6} \selectfont0.0018}           & 0.4008 $\pm$ \textbf{\fontsize{5}{6} \selectfont0.0030}       &0.7955 $\pm$ \textbf{\fontsize{5}{6} \selectfont0.0023}          & 0.04100 $\pm$ \textbf{\fontsize{5}{6} \selectfont0.0012}      & 0.9799 $\pm$ \textbf{\fontsize{5}{6} \selectfont0.0014}      & 0.8809 $\pm$ \textbf{\fontsize{5}{6} \selectfont0.0015}                   \\
				
				&  LDL-LRR     & 0.1122 $\pm$ \textbf{\fontsize{5}{6} \selectfont0.0030}           & 0.4772 $\pm$ \textbf{\fontsize{5}{6} \selectfont0.0036}      &0.8802 $\pm$ \textbf{\fontsize{5}{6} \selectfont0.0024}          & 0.05533 $\pm$ \textbf{\fontsize{5}{6} \selectfont0.0049}     & 0.9510 $\pm$ \textbf{\fontsize{5}{6} \selectfont0.0022}      & 0.8555 $\pm$ \textbf{\fontsize{5}{6} \selectfont0.0047}                       \\
				
				&  LDL-LCLR    & 0.1057 $\pm$ \textbf{\fontsize{5}{6} \selectfont0.0019}           & 1.0569 $\pm$ \textbf{\fontsize{5}{6} \selectfont0.0039}       &0.7890 $\pm$ \textbf{\fontsize{5}{6} \selectfont0.0039}          & 0.05045 $\pm$ \textbf{\fontsize{5}{6} \selectfont0.0037}     & 0.9668 $\pm$ \textbf{\fontsize{5}{6} \selectfont0.0049}      & 0.8383 $\pm$ \textbf{\fontsize{5}{6} \selectfont0.0018}                      \\
				
				&  LDLSF       & 0.1009 $\pm$ \textbf{\fontsize{5}{6} \selectfont0.0038}           & 0.4199 $\pm$ \textbf{\fontsize{5}{6} \selectfont0.0044}       &0.9008 $\pm$ \textbf{\fontsize{5}{6} \selectfont0.0015}          & 0.05199 $\pm$ \textbf{\fontsize{5}{6} \selectfont0.0040}     & 0.9779 $\pm$ \textbf{\fontsize{5}{6} \selectfont0.0018}      & 0.8660 $\pm$ \textbf{\fontsize{5}{6} \selectfont0.0022}                       \\
				
				&  LALOT       & 0.0989 $\pm$ \textbf{\fontsize{5}{6} \selectfont0.0019}           & 0.6689 $\pm$ \textbf{\fontsize{5}{6} \selectfont0.0019}       &0.8089 $\pm$ \textbf{\fontsize{5}{6} \selectfont0.0049}          & 0.04778 $\pm$ \textbf{\fontsize{5}{6} \selectfont0.0018}     & 0.9476 $\pm$ \textbf{\fontsize{5}{6} \selectfont0.0020}      & 0.8700 $\pm$ \textbf{\fontsize{5}{6} \selectfont0.0033}                      \\
				
				\multirow{-6}{*}{SJAFFE} 		            &  BFGS-LLD    & 0.1229 $\pm$ \textbf{\fontsize{5}{6} \selectfont0.0039}           & 1.5657 $\pm$ \textbf{\fontsize{5}{6} \selectfont0.0021}       &0.7998 $\pm$ \textbf{\fontsize{5}{6} \selectfont0.0020}          & 0.04998 $\pm$ \textbf{\fontsize{5}{6} \selectfont0.0051}     & 0.9711 $\pm$ \textbf{\fontsize{5}{6} \selectfont0.0036}      & 0.8611 $\pm$ \textbf{\fontsize{5}{6} \selectfont0.0016}                       \\
				
				\midrule
				\rowcolor{defaultcolor} 
				&  Ours        & 0.0833 $\pm$ \textbf{\fontsize{5}{6} \selectfont0.0020}           & 0.3994 $\pm$ \textbf{\fontsize{5}{6} \selectfont0.0010}       &0.7611 $\pm$ \textbf{\fontsize{5}{6} \selectfont0.0020}          & 0.03650 $\pm$ \textbf{\fontsize{5}{6} \selectfont0.0014}      & 0.9811 $\pm$ \textbf{\fontsize{5}{6} \selectfont0.0015}      & 0.8900 $\pm$ \textbf{\fontsize{5}{6} \selectfont0.0017}                    \\
				
				&  LDL-LRR     & 0.1109 $\pm$ \textbf{\fontsize{5}{6} \selectfont0.0036}           & 0.4477 $\pm$ \textbf{\fontsize{5}{6} \selectfont0.0039}      &0.8666 $\pm$ \textbf{\fontsize{5}{6} \selectfont0.0026}          & 0.05344 $\pm$ \textbf{\fontsize{5}{6} \selectfont0.0028}     & 0.9597 $\pm$ \textbf{\fontsize{5}{6} \selectfont0.0029}      & 0.8592 $\pm$ \textbf{\fontsize{5}{6} \selectfont0.0033}                    \\
				
				&  LDL-LCLR    & 0.1100 $\pm$ \textbf{\fontsize{5}{6} \selectfont0.0025}           & 0.9660 $\pm$ \textbf{\fontsize{5}{6} \selectfont0.0039}       &0.7897 $\pm$ \textbf{\fontsize{5}{6} \selectfont0.0033}          & 0.05101 $\pm$ \textbf{\fontsize{5}{6} \selectfont0.0021}     & 0.9677 $\pm$ \textbf{\fontsize{5}{6} \selectfont0.0056}      & 0.8555 $\pm$ \textbf{\fontsize{5}{6} \selectfont0.0032}                   \\
				
				&  LDLSF       & 0.1009 $\pm$ \textbf{\fontsize{5}{6} \selectfont0.0038}           & 0.4199 $\pm$ \textbf{\fontsize{5}{6} \selectfont0.0044}       &0.9008 $\pm$ \textbf{\fontsize{5}{6} \selectfont0.0015}          & 0.05199 $\pm$ \textbf{\fontsize{5}{6} \selectfont0.0040}     & 0.9780 $\pm$ \textbf{\fontsize{5}{6} \selectfont0.0029}      & 0.8660 $\pm$ \textbf{\fontsize{5}{6} \selectfont0.0022}                     \\
				
				&  LALOT       & 0.0989 $\pm$ \textbf{\fontsize{5}{6} \selectfont0.0019}           & 0.6689 $\pm$ \textbf{\fontsize{5}{6} \selectfont0.0019}       &0.8089 $\pm$ \textbf{\fontsize{5}{6} \selectfont0.0049}          & 0.04778 $\pm$ \textbf{\fontsize{5}{6} \selectfont0.0018}     & 0.9476 $\pm$ \textbf{\fontsize{5}{6} \selectfont0.0020}      & 0.8700 $\pm$ \textbf{\fontsize{5}{6} \selectfont0.0033}                    \\
				
				\multirow{-6}{*}{SBU} 		            &  BFGS-LLD    & 0.1119 $\pm$ \textbf{\fontsize{5}{6} \selectfont0.0030}           & 1.4657 $\pm$ \textbf{\fontsize{5}{6} \selectfont0.0022}       &0.7700 $\pm$ \textbf{\fontsize{5}{6} \selectfont0.0025}          & 0.04932 $\pm$ \textbf{\fontsize{5}{6} \selectfont0.0053}     & 0.9753 $\pm$ \textbf{\fontsize{5}{6} \selectfont0.0036}      & 0.8710 $\pm$ \textbf{\fontsize{5}{6} \selectfont0.0019}                    \\
				
				\midrule
				\rowcolor{defaultcolor} 
				&  Ours        & 0.2998 $\pm$ \textbf{\fontsize{5}{6} \selectfont0.0020}           & 2.3374 $\pm$ \textbf{\fontsize{5}{6} \selectfont0.0018}       &6.5163 $\pm$ \textbf{\fontsize{5}{6} \selectfont0.0018}          & 0.8111 $\pm$ \textbf{\fontsize{5}{6} \selectfont0.0029}      & 0.7890 $\pm$ \textbf{\fontsize{5}{6} \selectfont0.0049}      & 0.5691 $\pm$ \textbf{\fontsize{5}{6} \selectfont0.0010}                    \\
				
				&  LDL-LRR     & 0.3889 $\pm$ \textbf{\fontsize{5}{6} \selectfont0.0111}           & 3.1698 $\pm$ \textbf{\fontsize{5}{6} \selectfont0.0031}      &6.8777 $\pm$ \textbf{\fontsize{5}{6} \selectfont0.0025}          & 0.8999 $\pm$ \textbf{\fontsize{5}{6} \selectfont0.0069}     & 0.7044 $\pm$ \textbf{\fontsize{5}{6} \selectfont0.0077}      & 0.5444 $\pm$ \textbf{\fontsize{5}{6} \selectfont0.0049}                  \\
				
				&  LDL-LCLR    & 0.3740 $\pm$ \textbf{\fontsize{5}{6} \selectfont0.0066}           & 2.4986 $\pm$ \textbf{\fontsize{5}{6} \selectfont0.0066}       &6.8600 $\pm$ \textbf{\fontsize{5}{6} \selectfont0.0067}          & 0.8559 $\pm$ \textbf{\fontsize{5}{6} \selectfont0.0039}     & 0.7119 $\pm$ \textbf{\fontsize{5}{6} \selectfont0.0122}      & 0.5119 $\pm$ \textbf{\fontsize{5}{6} \selectfont0.0081}                   \\
				
				&  LDLSF       & 0.3441 $\pm$ \textbf{\fontsize{5}{6} \selectfont0.0249}           & 2.9884 $\pm$ \textbf{\fontsize{5}{6} \selectfont0.0055}       &6.6900 $\pm$ \textbf{\fontsize{5}{6} \selectfont0.0055}          & 0.8391 $\pm$ \textbf{\fontsize{5}{6} \selectfont0.0044}     & 0.7336 $\pm$ \textbf{\fontsize{5}{6} \selectfont0.0088}      & 0.8660 $\pm$ \textbf{\fontsize{5}{6} \selectfont0.0041}                     \\
				
				&  LALOT       & 0.3129 $\pm$ \textbf{\fontsize{5}{6} \selectfont0.0152}           & 2.3999 $\pm$ \textbf{\fontsize{5}{6} \selectfont0.0044}       &6.6666 $\pm$ \textbf{\fontsize{5}{6} \selectfont0.0078}          & 0.8226 $\pm$ \textbf{\fontsize{5}{6} \selectfont0.0033}     & 0.7390 $\pm$ \textbf{\fontsize{5}{6} \selectfont0.0100}      & 0.5224 $\pm$ \textbf{\fontsize{5}{6} \selectfont0.0066}                   \\
				
				\multirow{-6}{*}{Scene} 		            &  BFGS-LLD    & 0.3598 $\pm$ \textbf{\fontsize{5}{6} \selectfont0.0020}           & 2.4998 $\pm$ \textbf{\fontsize{5}{6} \selectfont0.0033}       &6.7999 $\pm$ \textbf{\fontsize{5}{6} \selectfont0.0049}          & 0.8400 $\pm$ \textbf{\fontsize{5}{6} \selectfont0.0033}     & 0.7333 $\pm$ \textbf{\fontsize{5}{6} \selectfont0.0064}      & 0.5199 $\pm$ \textbf{\fontsize{5}{6} \selectfont0.0055}                    \\
				
				\midrule
				\rowcolor{defaultcolor} 
				&  Ours        & 0.0488 $\pm$ \textbf{\fontsize{5}{6} \selectfont0.0012}           & 2.1029 $\pm$ \textbf{\fontsize{5}{6} \selectfont0.0259}       &14.0888 $\pm$ \textbf{\fontsize{5}{6} \selectfont0.0551}          & 0.2335 $\pm$ \textbf{\fontsize{5}{6} \selectfont0.0044}      & 0.8395 $\pm$ \textbf{\fontsize{5}{6} \selectfont0.0032}      & 0.7984 $\pm$ \textbf{\fontsize{5}{6} \selectfont0.0066}                   \\
				
				&  LDL-LRR     & 0.0537 $\pm$ \textbf{\fontsize{5}{6} \selectfont0.0039}           & 2.2887 $\pm$ \textbf{\fontsize{5}{6} \selectfont0.0860}      &14.3550 $\pm$ \textbf{\fontsize{5}{6} \selectfont0.0144}          & 0.2559 $\pm$ \textbf{\fontsize{5}{6} \selectfont0.0077}     & 0.8288 $\pm$ \textbf{\fontsize{5}{6} \selectfont0.0144}      & 0.7789 $\pm$ \textbf{\fontsize{5}{6} \selectfont0.0040}                     \\
				
				&  LDL-LCLR    & 0.0511 $\pm$ \textbf{\fontsize{5}{6} \selectfont0.0022}           & 2.2201 $\pm$ \textbf{\fontsize{5}{6} \selectfont0.0444}       &14.2101 $\pm$ \textbf{\fontsize{5}{6} \selectfont0.0510}          & 0.2566 $\pm$ \textbf{\fontsize{5}{6} \selectfont0.0047}     & 0.8302 $\pm$ \textbf{\fontsize{5}{6} \selectfont0.0012}      & 0.7722 $\pm$ \textbf{\fontsize{5}{6} \selectfont0.0060}                    \\
				
				&  LDLSF       & 0.0513 $\pm$ \textbf{\fontsize{5}{6} \selectfont0.0030}           & 2.2221$\pm$ \textbf{\fontsize{5}{6} \selectfont0.0036}       &14.3667 $\pm$ \textbf{\fontsize{5}{6} \selectfont0.0265}          & 0.2445 $\pm$ \textbf{\fontsize{5}{6} \selectfont0.0077}     & 0.8320 $\pm$ \textbf{\fontsize{5}{6} \selectfont0.0010}      & 0.7701 $\pm$ \textbf{\fontsize{5}{6} \selectfont0.0026}                      \\
				
				&  LALOT       & 0.0505 $\pm$ \textbf{\fontsize{5}{6} \selectfont0.0033}           & 2.1989 $\pm$ \textbf{\fontsize{5}{6} \selectfont0.0194}       &14.1855 $\pm$ \textbf{\fontsize{5}{6} \selectfont0.0922}          & 0.2443 $\pm$ \textbf{\fontsize{5}{6} \selectfont0.0088}     & 0.8297 $\pm$ \textbf{\fontsize{5}{6} \selectfont0.0060}      & 0.7888 $\pm$ \textbf{\fontsize{5}{6} \selectfont0.0013}                     \\
				
				\multirow{-6}{*}{Gene} 		            &  BFGS-LLD    & 0.0578 $\pm$ \textbf{\fontsize{5}{6} \selectfont0.0066}           & 2.3008 $\pm$ \textbf{\fontsize{5}{6} \selectfont0.0188}       &14.3559 $\pm$ \textbf{\fontsize{5}{6} \selectfont0.1556}          & 0.2480 $\pm$ \textbf{\fontsize{5}{6} \selectfont0.0015}     & 0.8300$\pm$ \textbf{\fontsize{5}{6} \selectfont0.0049}      & 0.7786 $\pm$ \textbf{\fontsize{5}{6} \selectfont0.0070}                     \\
				
				\midrule
				\rowcolor{defaultcolor}
				&  Ours        & 0.1089 $\pm$ \textbf{\fontsize{5}{6} \selectfont0.0018}           & 0.5001 $\pm$ \textbf{\fontsize{5}{6} \selectfont0.0044}       &0.9722 $\pm$ \textbf{\fontsize{5}{6} \selectfont0.0040}          & 0.0977 $\pm$ \textbf{\fontsize{5}{6} \selectfont0.0008}      & 0.9485 $\pm$ \textbf{\fontsize{5}{6} \selectfont0.0061}      & 0.8602 $\pm$ \textbf{\fontsize{5}{6} \selectfont0.0006}                     \\
				
				&  LDL-LRR     & 0.1135 $\pm$ \textbf{\fontsize{5}{6} \selectfont0.0009}           & 0.5244$\pm$ \textbf{\fontsize{5}{6} \selectfont0.0010}      &1.1551 $\pm$ \textbf{\fontsize{5}{6} \selectfont0.0061}          & 0.1445 $\pm$ \textbf{\fontsize{5}{6} \selectfont0.0049}     & 0.9510 $\pm$ \textbf{\fontsize{5}{6} \selectfont0.0022}      & 0.8772 $\pm$ \textbf{\fontsize{5}{6} \selectfont0.0007}                    \\
				
				&  LDL-LCLR    & 0.1177 $\pm$ \textbf{\fontsize{5}{6} \selectfont0.0086}           & 0.5345 $\pm$ \textbf{\fontsize{5}{6} \selectfont0.0040}       &1.1533 $\pm$ \textbf{\fontsize{5}{6} \selectfont0.0111}          & 0.1559 $\pm$ \textbf{\fontsize{5}{6} \selectfont0.0030}     & 0.9360 $\pm$ \textbf{\fontsize{5}{6} \selectfont0.0049}      & 0.8222 $\pm$ \textbf{\fontsize{5}{6} \selectfont0.0011}                    \\
				
				&  LDLSF       & 0.1155 $\pm$ \textbf{\fontsize{5}{6} \selectfont0.0045}           & 0.5339 $\pm$ \textbf{\fontsize{5}{6} \selectfont0.0062}       &1.1152$\pm$ \textbf{\fontsize{5}{6} \selectfont0.0050}          & 0.1540 $\pm$ \textbf{\fontsize{5}{6} \selectfont0.0041}     & 0.9445 $\pm$ \textbf{\fontsize{5}{6} \selectfont0.0020}      & 0.8551 $\pm$ \textbf{\fontsize{5}{6} \selectfont0.0044}                     \\
				
				&  LALOT       & 0.1221 $\pm$ \textbf{\fontsize{5}{6} \selectfont0.0110}           & 0.5440 $\pm$ \textbf{\fontsize{5}{6} \selectfont0.0033}       &1.111 $\pm$ \textbf{\fontsize{5}{6} \selectfont0.0040}          & 0.1503 $\pm$ \textbf{\fontsize{5}{6} \selectfont0.0008}     & 0.9477 $\pm$ \textbf{\fontsize{5}{6} \selectfont0.0022}      & 0.8559 $\pm$ \textbf{\fontsize{5}{6} \selectfont0.0002}                    \\
				
				\multirow{-6}{*}{Movie} 		            &  BFGS-LLD    & 0.1310 $\pm$ \textbf{\fontsize{5}{6} \selectfont0.0032}           & 0.5230 $\pm$ \textbf{\fontsize{5}{6} \selectfont0.0022}       &1.1170 $\pm$ \textbf{\fontsize{5}{6} \selectfont0.0024}          & 0.1595 $\pm$ \textbf{\fontsize{5}{6} \selectfont0.0155}     & 0.9400 $\pm$ \textbf{\fontsize{5}{6} \selectfont0.0003}      & 0.8491 $\pm$ \textbf{\fontsize{5}{6} \selectfont0.0018}                    \\
				
				\midrule
				\rowcolor{defaultcolor}
				&  Ours        & 0.3763 $\pm$ \textbf{\fontsize{5}{6} \selectfont0.0022}           & 1.1560 $\pm$ \textbf{\fontsize{5}{6} \selectfont0.0102}       &2.0889 $\pm$ \textbf{\fontsize{5}{6} \selectfont0.0055}          & 0.4880 $\pm$ \textbf{\fontsize{5}{6} \selectfont0.0023}      & 0.7998 $\pm$ \textbf{\fontsize{5}{6} \selectfont0.0022}      & 0.6703 $\pm$ \textbf{\fontsize{5}{6} \selectfont0.0033}                   \\
				
				&  LDL-LRR     & 0.3993 $\pm$ \textbf{\fontsize{5}{6} \selectfont0.0010}           & 1.4990 $\pm$ \textbf{\fontsize{5}{6} \selectfont0.0166}      &2.1884 $\pm$ \textbf{\fontsize{5}{6} \selectfont0.0034}          & 0.5246 $\pm$ \textbf{\fontsize{5}{6} \selectfont0.0006}     & 0.7531 $\pm$ \textbf{\fontsize{5}{6} \selectfont0.0023}      & 0.6334 $\pm$ \textbf{\fontsize{5}{6} \selectfont0.0077}                     \\
				
				&  LDL-LCLR    & 0.4040 $\pm$ \textbf{\fontsize{5}{6} \selectfont0.0082}           & 1.2444 $\pm$ \textbf{\fontsize{5}{6} \selectfont0.0045}       &2.2000 $\pm$ \textbf{\fontsize{5}{6} \selectfont0.0009}          & 0.4996 $\pm$ \textbf{\fontsize{5}{6} \selectfont0.0013}     & 0.7760 $\pm$ \textbf{\fontsize{5}{6} \selectfont0.0079}      & 0.6555 $\pm$ \textbf{\fontsize{5}{6} \selectfont0.0012}                 \\
				
				&  LDLSF       & 0.4159 $\pm$ \textbf{\fontsize{5}{6} \selectfont0.0055}          & 1.3105 $\pm$ \textbf{\fontsize{5}{6} \selectfont0.0041}       &2.2155 $\pm$ \textbf{\fontsize{5}{6} \selectfont0.0076}          & 0.5002 $\pm$ \textbf{\fontsize{5}{6} \selectfont0.0006}     & 0.7552 $\pm$ \textbf{\fontsize{5}{6} \selectfont0.0004}      & 0.6234 $\pm$ \textbf{\fontsize{5}{6} \selectfont0.0033}                    \\
				
				&  LALOT       & 0.3881 $\pm$ \textbf{\fontsize{5}{6} \selectfont0.0099}           & 1.4883 $\pm$ \textbf{\fontsize{5}{6} \selectfont0.0012}       &2.1257 $\pm$ \textbf{\fontsize{5}{6} \selectfont0.0268}          & 0.4990 $\pm$ \textbf{\fontsize{5}{6} \selectfont0.0008}     & 0.7549 $\pm$ \textbf{\fontsize{5}{6} \selectfont0.0021}      & 0.6620 $\pm$ \textbf{\fontsize{5}{6} \selectfont0.0053}                    \\
				
				\multirow{-6}{*}{M2B} 		            &  BFGS-LLD    & 0.3811 $\pm$ \textbf{\fontsize{5}{6} \selectfont0.0044}          & 1.3650 $\pm$ \textbf{\fontsize{5}{6} \selectfont0.0002}       &2.1992 $\pm$ \textbf{\fontsize{5}{6} \selectfont0.0095}          & 0.4995 $\pm$ \textbf{\fontsize{5}{6} \selectfont0.0005}     & 0.7699 $\pm$ \textbf{\fontsize{5}{6} \selectfont0.0040}      & 0.6532 $\pm$ \textbf{\fontsize{5}{6} \selectfont0.0009}                   \\
				
				\midrule
				\rowcolor{defaultcolor}
				&  Ours        & 0.3895 $\pm$ \textbf{\fontsize{5}{6} \selectfont0.0021}           & 1.2640 $\pm$ \textbf{\fontsize{5}{6} \selectfont0.0111}       &2.1995 $\pm$ \textbf{\fontsize{5}{6} \selectfont0.0095}          & 0.4911 $\pm$ \textbf{\fontsize{5}{6} \selectfont0.0030}      & 0.6990 $\pm$ \textbf{\fontsize{5}{6} \selectfont0.0002}      & 0.6004 $\pm$ \textbf{\fontsize{5}{6} \selectfont0.0001}                   \\
				
				&  LDL-LRR     & 0.4159 $\pm$ \textbf{\fontsize{5}{6} \selectfont0.0010}           & 1.6680 $\pm$ \textbf{\fontsize{5}{6} \selectfont0.0122}      &2.2006 $\pm$ \textbf{\fontsize{5}{6} \selectfont0.0039}          & 0.5388 $\pm$ \textbf{\fontsize{5}{6} \selectfont0.0006}     & 0.6531 $\pm$ \textbf{\fontsize{5}{6} \selectfont0.0023}      & 0.5804 $\pm$ \textbf{\fontsize{5}{6} \selectfont0.0007}                  \\
				
				&  LDL-LCLR    & 0.4240 $\pm$ \textbf{\fontsize{5}{6} \selectfont0.0042}           & 1.3444 $\pm$ \textbf{\fontsize{5}{6} \selectfont0.0055}       &2.2450$\pm$ \textbf{\fontsize{5}{6} \selectfont0.0016}          & 0.5131 $\pm$ \textbf{\fontsize{5}{6} \selectfont0.0022}     & 0.6261 $\pm$ \textbf{\fontsize{5}{6} \selectfont0.0005}      & 0.5500 $\pm$ \textbf{\fontsize{5}{6} \selectfont0.0012}                  \\
				
				&  LDLSF       & 0.4360 $\pm$ \textbf{\fontsize{5}{6} \selectfont0.0015}           & 1.2185 $\pm$ \textbf{\fontsize{5}{6} \selectfont0.0022}       &2.2159 $\pm$ \textbf{\fontsize{5}{6} \selectfont0.0076}          & 0.5120 $\pm$ \textbf{\fontsize{5}{6} \selectfont0.0006}     & 0.6261 $\pm$ \textbf{\fontsize{5}{6} \selectfont0.0004}      & 0.5534 $\pm$ \textbf{\fontsize{5}{6} \selectfont0.0030}                     \\
				
				&  LALOT       & 0.3999 $\pm$ \textbf{\fontsize{5}{6} \selectfont0.0009}           & 1.4983 $\pm$ \textbf{\fontsize{5}{6} \selectfont0.0012}       &2.22007 $\pm$ \textbf{\fontsize{5}{6} \selectfont0.0158}          & 0.4995 $\pm$ \textbf{\fontsize{5}{6} \selectfont0.0002}     & 0.6549 $\pm$ \textbf{\fontsize{5}{6} \selectfont0.0020}      & 0.6411 $\pm$ \textbf{\fontsize{5}{6} \selectfont0.0044}                    \\
				
				\multirow{-6}{*}{SCUT} 		            &  BFGS-LLD    & 0.3992 $\pm$ \textbf{\fontsize{5}{6} \selectfont0.0055}          & 1.5656 $\pm$ \textbf{\fontsize{5}{6} \selectfont0.0163}       &2.2832 $\pm$ \textbf{\fontsize{5}{6} \selectfont0.0080}          & 0.4966 $\pm$ \textbf{\fontsize{5}{6} \selectfont0.0011}     & 0.6491 $\pm$ \textbf{\fontsize{5}{6} \selectfont0.0040}      & 0.6333 $\pm$ \textbf{\fontsize{5}{6} \selectfont0.0013}                    \\
				
				\midrule
				\rowcolor{defaultcolor}
				&  Ours        & 0.1251 $\pm$ \textbf{\fontsize{5}{6} \selectfont0.0002}           & 1.1890 $\pm$ \textbf{\fontsize{5}{6} \selectfont0.0120}       &2.0980  $\pm$ \textbf{\fontsize{5}{6} \selectfont0.0223}          & 0.1053 $\pm$ \textbf{\fontsize{5}{6} \selectfont0.0009}      & 0.9643 $\pm$ \textbf{\fontsize{5}{6} \selectfont0.0015}      & 0.8501 $\pm$ \textbf{\fontsize{5}{6} \selectfont0.0025}                 \\
				
				&  LDL-LRR     & 0.1313 $\pm$ \textbf{\fontsize{5}{6} \selectfont0.0031}           & 1.2519 $\pm$ \textbf{\fontsize{5}{6} \selectfont0.0038}      &2.1992 $\pm$ \textbf{\fontsize{5}{6} \selectfont0.0095}          & 0.1127 $\pm$ \textbf{\fontsize{5}{6} \selectfont0.0077}     & 0.9533 $\pm$ \textbf{\fontsize{5}{6} \selectfont0.0021}      & 0.8412 $\pm$ \textbf{\fontsize{5}{6} \selectfont0.0066}                  \\
				
				&  LDL-LCLR    & 0.1277 $\pm$ \textbf{\fontsize{5}{6} \selectfont0.0016}          & 1.1969 $\pm$ \textbf{\fontsize{5}{6} \selectfont0.0039}       &2.1194 $\pm$ \textbf{\fontsize{5}{6} \selectfont0.0046}          & 0.1135 $\pm$ \textbf{\fontsize{5}{6} \selectfont0.0006}     & 0.9588 $\pm$ \textbf{\fontsize{5}{6} \selectfont0.0044}      & 0.8483 $\pm$ \textbf{\fontsize{5}{6} \selectfont0.0014}                  \\
				
				&  LDLSF       & 0.1270 $\pm$ \textbf{\fontsize{5}{6} \selectfont0.0028}           & 1.1909 $\pm$ \textbf{\fontsize{5}{6} \selectfont0.0164}       &2.1846 $\pm$ \textbf{\fontsize{5}{6} \selectfont0.0119}          & 0.1193 $\pm$ \textbf{\fontsize{5}{6} \selectfont0.0041}     & 0.9609 $\pm$ \textbf{\fontsize{5}{6} \selectfont0.0019}      & 0.8460 $\pm$ \textbf{\fontsize{5}{6} \selectfont0.0007}                     \\
				
				&  LALOT       & 0.1306 $\pm$ \textbf{\fontsize{5}{6} \selectfont0.0022}           & 1.1921 $\pm$ \textbf{\fontsize{5}{6} \selectfont0.0015}       &2.1111 $\pm$ \textbf{\fontsize{5}{6} \selectfont0.0171}          & 0.1120 $\pm$ \textbf{\fontsize{5}{6} \selectfont0.0015}     & 0.9430 $\pm$ \textbf{\fontsize{5}{6} \selectfont0.0019}      & 0.8400 $\pm$ \textbf{\fontsize{5}{6} \selectfont0.0004}                   \\
				
				\multirow{-6}{*}{fbp5500} 		            &  BFGS-LLD    & 0.1299 $\pm$ \textbf{\fontsize{5}{6} \selectfont0.0049}           & 1.4655 $\pm$ \textbf{\fontsize{5}{6} \selectfont0.0041}       &2.1675 $\pm$ \textbf{\fontsize{5}{6} \selectfont0.0024}          & 0.1135 $\pm$ \textbf{\fontsize{5}{6} \selectfont0.0055}     & 0.9595 $\pm$ \textbf{\fontsize{5}{6} \selectfont0.0030}      & 0.8419 $\pm$ \textbf{\fontsize{5}{6} \selectfont0.0018}                   \\
				
				\midrule
				\rowcolor{defaultcolor}
				&  Ours        & 0.1456 $\pm$ \textbf{\fontsize{5}{6} \selectfont0.0021}           & 1.3651 $\pm$ \textbf{\fontsize{5}{6} \selectfont0.0441}       &2.6888 $\pm$ \textbf{\fontsize{5}{6} \selectfont0.0023}          & 0.2017 $\pm$ \textbf{\fontsize{5}{6} \selectfont0.0012}      & 0.9394 $\pm$ \textbf{\fontsize{5}{6} \selectfont0.0026}      & 0.8247 $\pm$ \textbf{\fontsize{5}{6} \selectfont0.0077}                   \\
				
				&  LDL-LRR     & 0.1526 $\pm$ \textbf{\fontsize{5}{6} \selectfont0.0033}           & 1.5651 $\pm$ \textbf{\fontsize{5}{6} \selectfont0.0111}      &2.7594 $\pm$ \textbf{\fontsize{5}{6} \selectfont0.0422}          & 0.2449 $\pm$ \textbf{\fontsize{5}{6} \selectfont0.0007}     & 0.9251 $\pm$ \textbf{\fontsize{5}{6} \selectfont0.0003}      & 0.8141 $\pm$ \textbf{\fontsize{5}{6} \selectfont0.0044}                      \\
				
				&  LDL-LCLR    & 0.1515 $\pm$ \textbf{\fontsize{5}{6} \selectfont0.0022}           & 1.592 $\pm$ \textbf{\fontsize{5}{6} \selectfont0.0117}       &2.7779 $\pm$ \textbf{\fontsize{5}{6} \selectfont0.0239}          & 0.2244 $\pm$ \textbf{\fontsize{5}{6} \selectfont0.0030}     & 0.9262 $\pm$ \textbf{\fontsize{5}{6} \selectfont0.0062}      & 0.8189 $\pm$ \textbf{\fontsize{5}{6} \selectfont0.0098}                   \\
				
				&  LDLSF       & 0.1488 $\pm$ \textbf{\fontsize{5}{6} \selectfont0.0024}           & 1.3889$\pm$ \textbf{\fontsize{5}{6} \selectfont0.0086}       &2.7672 $\pm$ \textbf{\fontsize{5}{6} \selectfont0.0660}          & 0.2302 $\pm$ \textbf{\fontsize{5}{6} \selectfont0.0044}     & 0.9111 $\pm$ \textbf{\fontsize{5}{6} \selectfont0.0051}      & 0.8117 $\pm$ \textbf{\fontsize{5}{6} \selectfont0.0022}                     \\
				
				&  LALOT       & 0.1479 $\pm$ \textbf{\fontsize{5}{6} \selectfont0.0010}           & 1.3659 $\pm$ \textbf{\fontsize{5}{6} \selectfont0.0099}       &2.6956 $\pm$ \textbf{\fontsize{5}{6} \selectfont0.0144}          & 0.2221 $\pm$ \textbf{\fontsize{5}{6} \selectfont0.0064}     & 0.9311 $\pm$ \textbf{\fontsize{5}{6} \selectfont0.0021}      & 0.8107 $\pm$ \textbf{\fontsize{5}{6} \selectfont0.0008}                    \\
				
				\multirow{-6}{*}{RAF-ML} 		            &  BFGS-LLD    & 0.1499 $\pm$ \textbf{\fontsize{5}{6} \selectfont0.0009}           & 1.6656$\pm$ \textbf{\fontsize{5}{6} \selectfont0.0066}       &2.7101 $\pm$ \textbf{\fontsize{5}{6} \selectfont0.0211}          & 0.2541 $\pm$ \textbf{\fontsize{5}{6} \selectfont0.0055}     & 0.9204 $\pm$ \textbf{\fontsize{5}{6} \selectfont0.0023}      & 0.8157 $\pm$ \textbf{\fontsize{5}{6} \selectfont0.0050}                    \\
				
				\midrule
				\rowcolor{defaultcolor}
				&  Ours        & 0.2777 $\pm$ \textbf{\fontsize{5}{6} \selectfont0.0021}           & 2.2374 $\pm$ \textbf{\fontsize{5}{6} \selectfont0.0110}       &5.1163 $\pm$ \textbf{\fontsize{5}{6} \selectfont0.0018}          & 0.5111 $\pm$ \textbf{\fontsize{5}{6} \selectfont0.0029}      & 0.8807 $\pm$ \textbf{\fontsize{5}{6} \selectfont0.0049}      & 0.7891 $\pm$ \textbf{\fontsize{5}{6} \selectfont0.0014}                  \\
				
				&  LDL-LRR     & 0.3129 $\pm$ \textbf{\fontsize{5}{6} \selectfont0.0021}           & 3.2441 $\pm$ \textbf{\fontsize{5}{6} \selectfont0.0031}      &6.1454 $\pm$ \textbf{\fontsize{5}{6} \selectfont0.0023}          & 0.6616 $\pm$ \textbf{\fontsize{5}{6} \selectfont0.0035}     & 0.8002 $\pm$ \textbf{\fontsize{5}{6} \selectfont0.0042}      & 0.7411 $\pm$ \textbf{\fontsize{5}{6} \selectfont0.0014}                     \\
				
				&  LDL-LCLR    & 0.2994$\pm$ \textbf{\fontsize{5}{6} \selectfont0.0045}           & 2.4900 $\pm$ \textbf{\fontsize{5}{6} \selectfont0.0012}       &6.9609 $\pm$ \textbf{\fontsize{5}{6} \selectfont0.0041}          & 0.6056 $\pm$ \textbf{\fontsize{5}{6} \selectfont0.0031}     & 0.7110 $\pm$ \textbf{\fontsize{5}{6} \selectfont0.0021}      & 0.7110 $\pm$ \textbf{\fontsize{5}{6} \selectfont0.0088}                     \\
				
				&  LDLSF       & 0.3007 $\pm$ \textbf{\fontsize{5}{6} \selectfont0.0002}           & 2.7887 $\pm$ \textbf{\fontsize{5}{6} \selectfont0.0057}       &5.6101 $\pm$ \textbf{\fontsize{5}{6} \selectfont0.0118}         & 0.6396 $\pm$ \textbf{\fontsize{5}{6} \selectfont0.0022}     & 0.7939 $\pm$ \textbf{\fontsize{5}{6} \selectfont0.0098}      & 0.7660 $\pm$ \textbf{\fontsize{5}{6} \selectfont0.0007}                       \\
				
				&  LALOT       & 0.3133 $\pm$ \textbf{\fontsize{5}{6} \selectfont0.0021}           & 2.3141 $\pm$ \textbf{\fontsize{5}{6} \selectfont0.0016}       &5.5336 $\pm$ \textbf{\fontsize{5}{6} \selectfont0.0241}          & 0.5233 $\pm$ \textbf{\fontsize{5}{6} \selectfont0.0012}     & 0.8595 $\pm$ \textbf{\fontsize{5}{6} \selectfont0.055}      & 0.7214 $\pm$ \textbf{\fontsize{5}{6} \selectfont0.0049}                    \\
				
				\multirow{-6}{*}{Twitter} 		            &  BFGS-LLD    & 0.3114 $\pm$ \textbf{\fontsize{5}{6} \selectfont0.0044}           & 2.5511 $\pm$ \textbf{\fontsize{5}{6} \selectfont0.0028}       &5.7145 $\pm$ \textbf{\fontsize{5}{6} \selectfont0.0041}          & 0.5461 $\pm$ \textbf{\fontsize{5}{6} \selectfont0.0153}     & 0.8335 $\pm$ \textbf{\fontsize{5}{6} \selectfont0.0055}      & 0.7744 $\pm$ \textbf{\fontsize{5}{6} \selectfont0.0020}     \\
				
				\midrule
				\rowcolor{defaultcolor}
				&  Ours        & 0.2816 $\pm$ \textbf{\fontsize{5}{6} \selectfont0.0031}           & 2.3356 $\pm$ \textbf{\fontsize{5}{6} \selectfont0.0097}       &5.2222 $\pm$ \textbf{\fontsize{5}{6} \selectfont0.0159}          & 0.5314 $\pm$ \textbf{\fontsize{5}{6} \selectfont0.0033}      & 0.8406 $\pm$ \textbf{\fontsize{5}{6} \selectfont0.0041}      & 0.7741 $\pm$ \textbf{\fontsize{5}{6} \selectfont0.0025}                 \\
				
				&  LDL-LRR     & 0.3329 $\pm$ \textbf{\fontsize{5}{6} \selectfont0.0012}           & 3.4400 $\pm$ \textbf{\fontsize{5}{6} \selectfont0.0174}      &6.3459 $\pm$ \textbf{\fontsize{5}{6} \selectfont0.0229}          & 0.6516 $\pm$ \textbf{\fontsize{5}{6} \selectfont0.0031}     & 0.8450$\pm$ \textbf{\fontsize{5}{6} \selectfont0.0040}      & 0.7399 $\pm$ \textbf{\fontsize{5}{6} \selectfont0.0037}                    \\
				
				&  LDL-LCLR    & 0.2970$\pm$ \textbf{\fontsize{5}{6} \selectfont0.0009}           & 2.4444 $\pm$ \textbf{\fontsize{5}{6} \selectfont0.0063}       &6.1600 $\pm$ \textbf{\fontsize{5}{6} \selectfont0.0041}         & 0.6222 $\pm$ \textbf{\fontsize{5}{6} \selectfont0.0013}     & 0.7919 $\pm$ \textbf{\fontsize{5}{6} \selectfont0.0029}      & 0.7090$\pm$ \textbf{\fontsize{5}{6} \selectfont0.0070}                     \\
				
				&  LDLSF       & 0.3301 $\pm$ \textbf{\fontsize{5}{6} \selectfont0.0009}           & 2.8888 $\pm$ \textbf{\fontsize{5}{6} \selectfont0.0459}       &5.9152 $\pm$ \textbf{\fontsize{5}{6} \selectfont0.0121}          & 0.6100 $\pm$ \textbf{\fontsize{5}{6} \selectfont0.0021}     & 0.8139 $\pm$ \textbf{\fontsize{5}{6} \selectfont0.0098}      & 0.7360 $\pm$ \textbf{\fontsize{5}{6} \selectfont0.0037}                      \\
				
				&  LALOT       & 0.3411 $\pm$ \textbf{\fontsize{5}{6} \selectfont0.0026}           & 2.9140$\pm$ \textbf{\fontsize{5}{6} \selectfont0.0019}       &5.3333 $\pm$ \textbf{\fontsize{5}{6} \selectfont0.0243}          & 0.5737 $\pm$ \textbf{\fontsize{5}{6} \selectfont0.0012}     & 0.8225 $\pm$ \textbf{\fontsize{5}{6} \selectfont0.020}     & 0.7144 $\pm$ \textbf{\fontsize{5}{6} \selectfont0.0004}                    \\
				
				\multirow{-6}{*}{Flickr} 		            &  BFGS-LLD    & 0.3200$\pm$ \textbf{\fontsize{5}{6} \selectfont0.0041}           & 2.7517 $\pm$ \textbf{\fontsize{5}{6} \selectfont0.0060}       &5.8149 $\pm$ \textbf{\fontsize{5}{6} \selectfont0.0048}          & 0.5961 $\pm$ \textbf{\fontsize{5}{6} \selectfont0.0099}     & 0.8131 $\pm$ \textbf{\fontsize{5}{6} \selectfont0.0011}      & 0.7407 $\pm$ \textbf{\fontsize{5}{6} \selectfont0.0077}       \\
				
				\bottomrule
			\end{tabular}%
%		}
		\vspace{-1mm}
	\end{center}
\end{table*}

\vspace{-4mm}
{\flushleft \textbf{Results and Discussion.}}
We conduct 10 times 5-fold cross-validation on each dataset. 
The experimental results are presented in the form of ``mean$\pm$std'' in Table~\ref{T3}.
Overall, our proposed method outperforms other comparison algorithms in all evaluation metrics.
Each comparison algorithm employs some regularization techniques to expand the training sample as well as to prevent overfitting, however, three main factors contribute to the competitive results of our approach.
\textbf{i):}  Moderate noise, especially on the Gene dataset, due to the uncertainty that comes with manual annotation, our approach has a huge performance gain with the help of implicit distribution representation with Gaussian priors.
\textbf{ii):} The ability to capture global features between labels with the help of a self-attentive mechanism. Also, LALOT performs sub-optimally probably because of global modeling. 
\textbf{iii):} The powerful representational capabilities of the deep network, especially on image datasets, give us a huge advantage.
%
%In addition, the analysis of variance (ANOVA) tests is conducted on each dataset to compare the equality of the LDL methods.
%
%The test results (ANOVA is mentioned in the supplementary material) reported significantly different performance of the LDL method for each metric on all datasets except Gene ( at a significance level of 0.05).
%
%The label distribution of Gene tends to be uniformly distributed, which may lead to the same performance as the linear-based LDL method.
%

\begin{table*}[!htb] \scriptsize
	\begin{center}
		\vspace{-0mm}
		\caption{Ablation study (AS). Effectiveness of the loss functions and the modules on two datasets. Quantitative results demonstrate the effectiveness of each module.}
		\vspace{-0mm}
		\label{T4}
		\resizebox{\linewidth}{!}{
			\begin{tabular}{c|c|cccccc|c}
				\toprule
				AS                             & Algorithm	   & Chebyshev $\downarrow$  & Clark $\downarrow$   &Canberra $\downarrow$   &K-L $\downarrow$    &Cosine $\uparrow$   &Intersection $\uparrow$         & Dataset   \\ 	
				\midrule                           
				\rowcolor{defaultcolor}
				&  Ours        & 0.0488 $\pm$ \textbf{\fontsize{5}{6} \selectfont0.0012}           & 2.1029 $\pm$ \textbf{\fontsize{5}{6} \selectfont0.0259}       &14.0888 $\pm$ \textbf{\fontsize{5}{6} \selectfont0.0551}          & 0.2335 $\pm$ \textbf{\fontsize{5}{6} \selectfont0.0044}      & 0.8395 $\pm$ \textbf{\fontsize{5}{6} \selectfont0.0032}      & 0.7984 $\pm$ \textbf{\fontsize{5}{6} \selectfont0.0066}                  \\
				
				\multirow{-2}{*}{\textcolor{black!98}{(a)}} 	&  w/o $\mathcal{L}_{p}$     & 0.0497 $\pm$ \textbf{\fontsize{5}{6} \selectfont0.0021}           & 2.1664 $\pm$ \textbf{\fontsize{5}{6} \selectfont0.0177}     &14.2651 $\pm$ \textbf{\fontsize{5}{6} \selectfont0.0155}          & 0.2488 $\pm$ \textbf{\fontsize{5}{6} \selectfont0.0071}     & 0.8290 $\pm$ \textbf{\fontsize{5}{6} \selectfont0.074}     & 0.7884 $\pm$ \textbf{\fontsize{5}{6} \selectfont0.0037}      &\multirow{-2}{*}{Gene}                \\

				\midrule
				\rowcolor{defaultcolor}
				&  Ours        & 0.0779 $\pm$ \textbf{\fontsize{5}{6} \selectfont0.0021}           & 0.3980 $\pm$ \textbf{\fontsize{5}{6} \selectfont0.0051}       &0.7779 $\pm$ \textbf{\fontsize{5}{6} \selectfont0.0030}          & 0.04040 $\pm$ \textbf{\fontsize{5}{6} \selectfont0.0020}      & 0.9883 $\pm$ \textbf{\fontsize{5}{6} \selectfont0.0009}      & 0.8778 $\pm$ \textbf{\fontsize{5}{6} \selectfont0.0014}                 \\
				
				&  w/o PRT        & 0.0877 $\pm$ \textbf{\fontsize{5}{6} \selectfont0.0009}           & 0.4008 $\pm$ \textbf{\fontsize{5}{6} \selectfont0.0043}       &0.7881 $\pm$ \textbf{\fontsize{5}{6} \selectfont0.0014}          & 0.04223 $\pm$ \textbf{\fontsize{5}{6} \selectfont0.0010}      & 0.9779 $\pm$ \textbf{\fontsize{5}{6} \selectfont0.0008}      & 0.8699 $\pm$ \textbf{\fontsize{5}{6} \selectfont0.0012}                 \\
				
				&  SNN        & 0.0771 $\pm$ \textbf{\fontsize{5}{6} \selectfont0.0012}           & 0.4006 $\pm$ \textbf{\fontsize{5}{6} \selectfont0.0047}       &0.7805 $\pm$ \textbf{\fontsize{5}{6} \selectfont0.0011}          & 0.04118 $\pm$ \textbf{\fontsize{5}{6} \selectfont0.0016}      & 0.9801 $\pm$ \textbf{\fontsize{5}{6} \selectfont0.0012}      & 0.8664 $\pm$ \textbf{\fontsize{5}{6} \selectfont0.0029}                 \\

				\multirow{-4}{*}{(\textcolor{black!98}{b},\textcolor{black!98}{c},\textcolor{black!98}{d})} 		            &  GNN    & 0.0804 $\pm$ \textbf{\fontsize{5}{6} \selectfont0.0033}           & 0.4133 $\pm$ \textbf{\fontsize{5}{6} \selectfont0.0017}       &0.7968 $\pm$ \textbf{\fontsize{5}{6} \selectfont0.0020}          & 0.04991 $\pm$ \textbf{\fontsize{5}{6} \selectfont0.0052}     &0.9705 $\pm$ \textbf{\fontsize{5}{6} \selectfont0.0036}      & 0.8612 $\pm$ \textbf{\fontsize{5}{6} \selectfont0.0016}      &\multirow{-4}{*}{wc-LDL}               \\

				\bottomrule
			\end{tabular}%
		}
		\vspace{-2mm}
	\end{center}
\end{table*}
\vspace{-4mm}
{\flushleft \textbf{Ablation Study.}}
\vspace{-0mm}
To demonstrate the effectiveness of the loss function and the module of our model, we conduct an ablation study involving the following four experiments:
\textbf{\textcolor{black!98}{(a)}} w/o perceptual loss function $\mathcal{L}_{p}$: we remove the loss function on training Gene dataset, shown in Table~\ref{T4}.
\textbf{\textcolor{black!98}{(b)}} w/o our proposed regularization techniques (PRT): we remove the linear normalization function and the data augmentation respectively on the training wc-LDL dataset, shown in Table~\ref{T4}. 
\textbf{\textcolor{black!98}{(c)}} The effectiveness of SNN: we use standard MLPs to replace SNNs in the same network architecture, shown in Table~\ref{T4}. 
\textbf{\textcolor{black!98}{(d)}} The effectiveness of GNN: for deep implicit function construction, we use standard MLPs to replace GNNs, as shown in Table~\ref{T4}. 
We conduct 10 times 5-fold cross-validation on the dataset of the ablation experiment. 

\vspace{-2mm}
{\flushleft \textbf{Potential of Model and Network.}}
Our model can be extended to handle classification and semi-supervised tasks.
We conduct some experiment reports to show the potential of the public datasets.
In addition, there is a deep model (DLDL (\cite{gao2017deep})) based on a label distribution learning framework as a training scheme being performed on several classification tasks.
Since DLDL does not have an adapted network implementation on tabular data, to demonstrate the effectiveness of our proposed method, our approach battles with the DLDL algorithm in the age estimation dataset.

\begin{figure*}[!htb]\scriptsize
	\begin{center}
		\vspace{-4mm}
		\tabcolsep 1pt
		\begin{tabular}{@{}ccc@{}}
			\includegraphics[width = 0.31\textwidth]{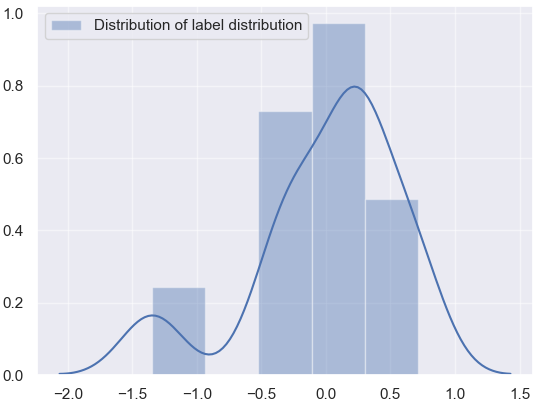}         &
			\includegraphics[width = 0.31\textwidth]{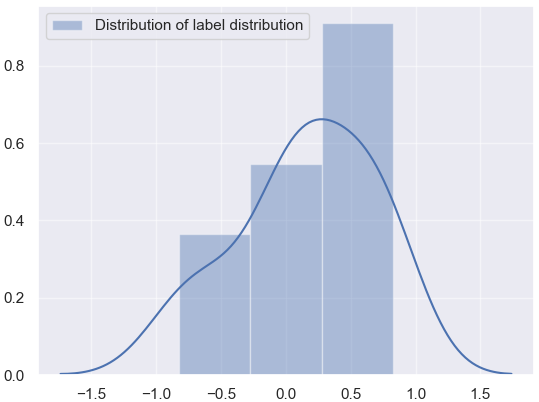}         &
			\includegraphics[width = 0.31\textwidth]{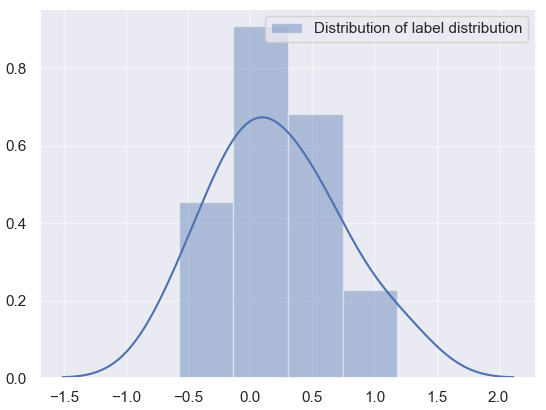}         \\
			(a) L1 of the matrix &
			(b) L2 of the matrix &
			(c) L3 of the matrix \\
			\includegraphics[width = 0.31\textwidth]{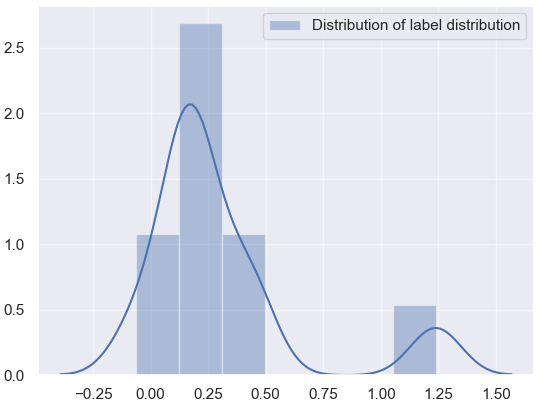}         &
			\includegraphics[width = 0.31\textwidth]{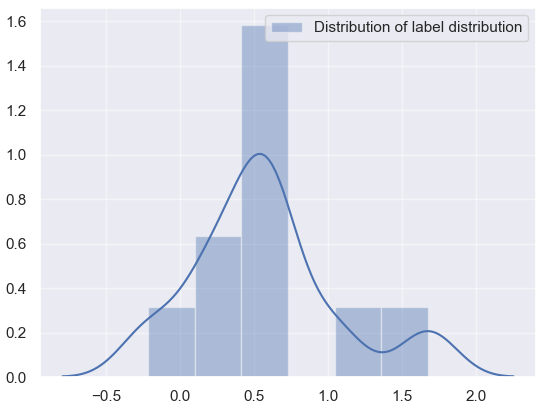}         &
			\includegraphics[width = 0.31\textwidth]{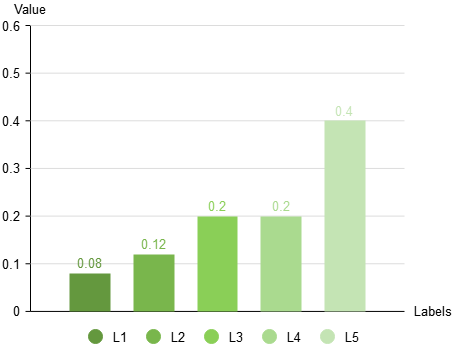}         \\       
			(d) L4 of the matrix &
			(e) L5 of the matrix &
			(f) Label distribution   \\
		\end{tabular}
	\end{center}
	\vspace{-2mm}
	\caption{This figure shows the distribution of data for each vector of the label distribution matrix and the corresponding label for this label distribution matrix.}
	\vspace{-2mm}
	\label{fig-3}
\end{figure*}

\vspace{-2mm}
(1) Facial expression recognition (\cite{zhao2021robust}):
To evaluate the effectiveness of the proposed model, we conduct the experiments on the public in-the-wild facial expression datasets (RAF-DB, CAER-S, AffectNet).
For images on all the datasets, the face region is detected and aligned by using Retinaface (\cite{deng2020retinaface}).
Then, the image is cropped to a fixed resolution (224 $\times$ 224) by bilinear interpolation.
Our approach is pre-trained on the face recognition dataset MS-Celeb-1M, and the 50-layer Residual Network is used as the backbone network.
For our network, parameters were optimized using the Adam optimizer with an initial learning rate of 0.01, a mini-batch size of 128, and an epoch is 50.
Notably, our network architecture uses a model (both input and output layers are changed in the number of neurons and the rest are fixed) executed on the Gene dataset with a single RTX GPU, and the loss function uses cross-entropy.
We compared it with the current SOTA model (Face2Exp) on three popular datasets.
The recognition accuracy of Face2Exp in the three data sets (RAF-DB, CAER-S, AffectNet) is 88.54\%, 86.16\%, and 64.23\% respectively.
The recognition accuracy of our proposed model in three datasets is 88.52\%, 86.36\%, and 65.02\% respectively.  
Our algorithm achieves competitive results on most of the face recognition datasets.
%
%Gene requires the prediction of a large number of labels, which introduces an effect of distillation (the role of $\tau$) on the label space.
(2) MedMNIST Classification Decathlon (\cite{yang2021medmnist}):
We evaluate the algorithm's performance on the MedMNIST Classification Decathlon benchmark.
The area under the ROC curve (AUC) and Accuracy (ACC) is used as the evaluation metrics.
Our approach is pre-trained on the Gene dataset, the 18-layer Residual Network is adopted as the backbone network.
Our model is trained for 100 epochs, using a cross-entropy loss and an Adam optimizer with a batch size of 128 and an initial learning rate of $1 \times 10 ^{-3}$
The overall performance of the methods is reported in Table~\ref{tab:Results}.
Our model is pre-trained during the training phase and thus achieves competitive performance on most of the datasets.
Literature (\cite{yang2021medmnist}) includes the bibliography of the full comparison method.
\begin{table*}[!htb] \tiny
	\caption{{\bf Overall performance of MedMNIST} in metrics of AUC and ACC, using ResNet-18 / ResNet-50 with resolution $28$ and $224$, auto-sklearn , AutoKeras and Google AutoML Vision.}
	\label{tab:Results}
	\vspace{-1mm}
	\begin{center}
		
		\begin{tabular}{@{}ccccccccccc@{}}
			\toprule
			\multirow{2}{*}{Methods} &
			\multicolumn{2}{c}{PathMNIST} &
			\multicolumn{2}{c}{ChestMNIST} &
			\multicolumn{2}{c}{DermaMNIST} &
			\multicolumn{2}{c}{OCTMNIST} &
			\multicolumn{2}{c}{PneumoniaMNIST} \\
			& AUC & ACC & AUC & ACC & AUC & ACC & AUC & ACC & AUC & ACC \\ \midrule
			ResNet-18 (28)     & 0.972 & 0.844 & 0.706 & 0.947 & 0.899 & 0.721 & 0.951 & \bf0.758 & 0.957 & 0.843  \\
			ResNet-18 (224)        & 0.978 & 0.860 & 0.713 & \bf0.948 & 0.896 & 0.727 & 0.960 & 0.752 & 0.970 & 0.861  \\
			ResNet-50 (28)        & 0.979 & \bf0.864 & 0.692 & 0.947 & 0.886 & 0.710 & 0.939 & 0.745 & 0.949 & 0.857  \\
			ResNet-50 (224)       & 0.978 & 0.848 & 0.706 & 0.947 & 0.895 & 0.719 & 0.951 & 0.750 & 0.968 & 0.896  \\
			auto-sklearn         & 0.500 & 0.186 & 0.647 & 0.642 & 0.906 & 0.734 & 0.883 & 0.595 & 0.947 & 0.865  \\
			AutoKeras           & 0.979 & \bf0.864 & 0.715 & 0.939 & 0.921 & 0.756 & 0.956 & 0.736 & 0.970 & 0.918 \\
			Google AutoML Vision  & 0.982 & 0.811 & 0.718 & 0.947 & 0.925 & \bf0.766 & 0.965 & 0.732 & \bf0.993 & 0.941 \\
			Ours  & \bf0.985& \bf0.864 & \bf0.722 & \bf0.948 & \bf0.929 & 0.765 & \bf0.969 & 0.756 & 0.992 & \bf0.943 \\
			\midrule
			\multirow{2}{*}{Methods} &
			\multicolumn{2}{c}{RetinaMNIST} &
			\multicolumn{2}{c}{BreastMNIST} &
			\multicolumn{2}{c}{OrganMNIST\_A} &
			\multicolumn{2}{c}{OrganMNIST\_C} &
			\multicolumn{2}{c}{OrganMNIST\_S} \\
			& AUC & ACC & AUC & ACC & AUC & ACC & AUC & ACC & AUC & ACC \\ \midrule
			ResNet-18 (28)         & 0.727 & 0.515 & 0.897 & 0.859 & 0.995 & 0.921 & 0.990 & 0.889 & 0.967 & 0.762 \\
			ResNet-18 (224)       & 0.721 & 0.543 & 0.915 & 0.878 & \bf0.997 & 0.931 & 0.991 & 0.907 & \bf0.974 & 0.777 \\
			ResNet-50 (28)         & 0.719 & 0.490 & 0.879 & 0.853 & 0.995 & 0.916 & 0.990 & 0.893 & 0.968 & 0.746 \\
			ResNet-50 (224)        & 0.717 & 0.555 & 0.863 & 0.833 & \bf0.997 & 0.931 & \bf0.992 & 0.898 & 0.970 & 0.770 \\
			auto-sklearn        & 0.694 & 0.525 & 0.848 & 0.808 & 0.797 & 0.563 & 0.898 & 0.676 & 0.855 & 0.601 \\
			AutoKeras           & 0.655 & 0.420 & 0.833 & 0.801 & 0.996 & 0.929 & \bf0.992 & 0.915 & 0.972 & 0.803 \\
			Google AutoML Vision  & \bf0.762 & 0.530    &  \bf0.932 & 0.865    & 0.988 & 0.818 & 0.986 & 0.861 & 0.964 & 0.706 \\
			Ours                  & 0.760    & \bf0.570 & 0.931     & \bf0.890 & 0.996 & \bf0.933 & 0.990 & \bf0.917 & 0.971 & \bf0.810 \\
			\bottomrule
		\end{tabular}
		\vspace{-2mm}
	\end{center}
\end{table*}
(3) CIFAR-10, SVHN, CIFAR-100 (\cite{hu2021simple}):
Since our approach involves graph structure, it battles with other SOTAs on the semi-supervised task.
We use ResNet-28-2 as our backbone and Adam with weight decay for optimization in all experiments.
Our model is trained for 100 epochs, using a cross-entropy loss with a batch size of 128 and an initial learning rate of $1 \times 10 ^{-2}$
Early-stopping and data augmentation (Mixup) are also adopted in the training phase.
Two datasets (CIFAR-10, and SVHN) are used for us to evaluate the performance of the algorithm.
The overall performance of the methods is reported in Table~\ref{warps}.
Our model still selects MLPs pre-trained on the Gene dataset and performs competitively on the semi-supervised task.
Literature (\cite{hu2021simple}) includes the bibliography of the full comparison method.
\begin{table*}[t] \tiny 
		\centering
		\caption{CIFAR-10, CIFAR-100, and SVHN Top-1 test accuracy.}
		\label{warps}
		\begin{tabular}{@{}ccccccc|cc@{}}
			\toprule
			\multirow{2}{*}{Methods} &
			\multicolumn{2}{c}{CIFAR-10} &
			\multicolumn{2}{c}{SVHN}  &
			\multicolumn{2}{c}{CIFAR-100}&
			\multirow{2}{*}{Backbone} \\
			 &1000 labels & 4000 labels & 1000 labels & 4000 labels  &10000 labels &15000 labels & \\ \midrule
		    VAT               & 81.36 & 88.95 & 94.02 & 95.80  &77.54 &81.55 & WRN-28-8\\
			MeanTeacher       & 82.68 & 89.64 & 96.25 & 96.61  &72.68 &79.62 & WRN-28-8\\
		    MixMatch         & 92.25 & 93.76 & 96.73 & 97.11  &71.69 &79.13 & WRN-28-8\\
			ReMixMatch        & 94.27 & 94.86 & 97.17 & 97.58  &76.97 &80.99 & WRN-28-8\\
			FixMatch~       & - & 95.69 & \bf97.64 & -    &77.40       &64.25 & WRN-28-8\\
			SimPLE           & 94.84 & 94.95 & 97.54 & 97.31  &78.11  &\bf82.40 & WRN-28-8\\
			Ours  							    & \bf94.98& \bf95.99 & 97.56 & \bf97.62  &\bf78.64 &81.02 & WRN-28-8\\
			\bottomrule
		\end{tabular}
\vspace{-2mm}
\end{table*}
(4) Semi-supervised label distribution learning (\cite{jia2021semi}):
To evaluate the capability of our model, our proposed algorithm is compared to the SOTA model (PGE-SLDL (\cite{jia2021semi})) in the Gene with a 50\% missing rate.
We conduct it 10 times on Gene and average the 10 results as the final result.
PGE-SLDL yield 0.0059$\pm$0.0008, 0.4298$\pm$0.0006, 4.3899$\pm$0.0005, 0.09788$\pm$0.0005, 0.8911$\pm$0.0008 and 0.8789$\pm$0.0007 respectively on the data set with 6 evaluation metrics (Chebyshev$\downarrow$, Clark$\downarrow$, Canberra$\downarrow$, K-L divergence$\downarrow$, Intersection$\uparrow$, Cosine$\uparrow$).
Our method yield 0.0053$\pm$0.0012, 0.4112$\pm$0.0005, 4.2990$\pm$0.0008, 0.09661$\pm$0.0005, 0.9001$\pm$0.0012 and 0.8797$\pm$0.0008 respectively on the data set with corresponding 6 evaluation metrics.
Our method still shows competitiveness, where the model remains fixed, the learning rate and the number of iterations are not changed, and the $\lambda_{2}$ of Equation 6 is adjusted to 0.25.
(5) Our approach vs. DLDL (Age estimation (\cite{gao2017deep})):
Two age estimation datasets are used in our experiments.
The first is Morph (\cite{ricanek2006morph}), which is one of the largest publicly available age datasets.
The second dataset is from the apparent age estimation competition, the first competition track of the ICCV ChaLearn LAP 2015 workshop (\cite{escalera2015chalearn}).
We employ the DPM model to detect the main facial region. 
Then, the detected face is fed into a cascaded convolution network to get the five facial key points, including the left/right eye centers, nose tip, and left/right mouth corners. 
Finally, based on these facial points, we align the face to the upright pose.
The implementation details of our method with the preprocessing unit are referenced in DLDL.
We used MAE and $\epsilon$-error to evaluate the performance of our model and the comparison method, respectively.
DLDL algorithm on Morph and ChaLearn with metrics MAE $\downarrow$ and $\epsilon$-error yields performance that is \{2.42, 0.23; 3.51, 0.31\}.
Our algorithm achieves competitive results (\textbf{\{2.40, 0.20; 3.23, 0.27\}}).
\vspace{-2mm}
{\flushleft \textbf{Visualize of Label Distribution Matrix.}} We conduct an experiment to verify the validity of our algorithm.
We extract one sample (including features and labeled distributions) on Movie, and then visualize the matrix of label distribution ($5 \times 10$) learned by the network and the corresponding label distribution.
As shown in the Figure~\ref{fig-3}, we notice that the data distribution (\textbf{means and variances}) of the label distribution vector (each row of the label distribution matrix) is consistent with the corresponding label distribution.

\vspace{-4mm}

\vspace{-4mm}
\section{Application and Energy Consumption}
\vspace{-3mm}
%
%Our model saves energy compared to standard deep networks while keeping accurate inference.
%
%This is thanks to the role played by the SNN model, i.e. discrete binary coding.
%
We select the algorithm executed on Gene as the evaluation model.
The evaluation model is conducted on three platforms (lynxi HP300, Raspberry Pi, and Apple smartphones) to check the energy consumption with the same number of iterations (the power is evaluated thanks to the adb script (\cite{dzhagaryan2016environment})).
Specifically, our model is trained on Gene with the accuracy of float16.
%
%After that, our model is compiled with the help of the conversion functions provided by spikingjelly.
%
Then, the baseline model (MLP replaces SNN and GNN) and our model run inference on a mobile platform, each model executing 500 epochs with 32 batch sizes in the iteration.
Experimental results show that our algorithm saves 34.6\%, 41.2\%, and 40.8\% energy compared to the baseline algorithm on the three platforms, respectively.
%
%Note that the feature data is provided by a random function.
%
%\vspace{-3mm}
%\section{Discussions and Social Impact.}
%\vspace{-3mm}
%
%We propose an algorithm that aims to create a novel learning framework to tackle the label distribution learning task rather than address the drawbacks of other models.
%
%Involving the potential of the network, by modeling on four tasks (2 classification tasks and 2 semi-supervised tasks) to show the competitiveness of the algorithm.
%In this paper, we propose a novel generator to tackle the label distribution learning task.
%
%Note that our proposed method is not evaluated on a large dataset (ImageNet) and does not show outstanding performance with a sufficient sample.
%
%This may be because GNNs with a small number of graph convolutions are not sufficient to build label relations in the open world.
%
%In the future, we will consider NAS as a tool to further boost model performance.
%
%New data set is collected and marshaled without privacy security issues.
%
%Evaluation on three mobile platforms is also compliant in terms of safety and ethics.
%
\vspace{-4mm}
\section{Related Works}
\vspace{-3mm}
{\flushleft \textbf{Label distribution Learning.}}
Label distribution learning has attracted several attention as a new learning paradigm.
Label distribution learning comes from the scheme proposed by (\cite{geng2016label}) to address the age estimation task.
Since then a large number of approaches have been proposed, such as low-rank hypothesis-based (\cite{jia2019facial,ren2019label}), metric-based (\cite{gao2018age}), manifold-based (\cite{wang2021label}), and label correlation-based (\cite{teng2021incomplete,qian2022feature}).
Among them, some approaches are executed in computer vision (\cite{chen2021toward}), and speech recognition (\cite{si2022towards}) tasks to improve the performance of classifiers.
In this paper, we try to build a distribution of label distributions to moderate noise and uncertainty.
%However, most of these approaches are based on discriminative learning patterns, and we propose a deep generative model to offer a new idea for tackling label distribution learning. 

\textbf{Implicit Neural Representations.}
% $F_{d} \in$ ${\mathbb{R}}^{N \times d}$
In implicit neural representation, an object is usually represented as a multi-layer perception (MLP) that maps coordinates to a signal.
This idea has been widely applied in modeling 3D object shapes (\cite{lin2020sdf,kohli2020inferring}), 3D surfaces of the scene (\cite{sitzmann2019scene,yariv2020multiview,niemeyer2020differentiable,jiang2020local}), the appearance of the 3D structure as well as the 2D image enhancement (\cite{skorokhodov2021adversarial,chen2021learning,anokhin2021image,karras2021alias}).
%
%Although these methods have achieved competitive performance, they have not yet been applied to low-dimensional signal prediction.
%
In this paper, we seek to explore this technique to address the label distribution learning issue.
\vspace{-4mm}
\section{Conclusion}
\vspace{-3mm}
In this paper, we design an implicit distribution representation algorithm to moderate the uncertainty of the label values, where the implicit function can be a good estimate of the continuous distribution space.
Furthermore, Gaussian prior methods and self-attention mechanisms help the model learn both local signals and global information of the label distribution matrix.
Numerous experiments have verified the effectiveness of our approach as well as the suitability of the regularization technique. 
The application session demonstrates the high efficiency of our proposed model.

\bibliography{iclr2023_conference}
\bibliographystyle{iclr2023_conference}
\end{document}